\input ./style/arxiv-general.cfg
\documentclass[MSNbibl,number,citesort,seceqn,dvips]{arxbj}
\makeatletter
   \@ifpackageloaded{graphicx}{}{\usepackage{graphicx}}
\makeatother
\usepackage{upgreek}

\volume{23}
\issue{1}
\pubyear{2017}
\firstpage{23}
\lastpage{57}
\doi{10.3150/15-BEJ702}
\docsubty{FLA}

\makeatletter
\def\pi{\uppi}

\newcommand{\rrvert}{\vert}

\newcommand{\rrVert}{\Vert}
\newcommand{\llvert}{\vert}
\newcommand{\llVert}{\Vert}
\newcommand{\eqref}[1]{(\ref{#1})}
\def\reals{{\mathbb R}}
\def\P{{\mathbb P}}
\def\E{{\mathbb E}}
\def\supp{\mathop{\operatorname{supp}}}
\def\S{\mathbb{S}}
\def\sign{\operatorname{sign}}
\def\card{\mathop{\operatorname{card}}}
\def\rank{\operatorname{rank}}
\def\vec{\mathop{\operatorname{vec}}}
\def\tr{\operatorname{Tr}}
\let\hat\widehat
\let\tilde\widetilde

\newcommand{\ba}{\mathbf{a}}
\newcommand{\bv}{\mathbf{v}}

\newcommand{\Ab}{\mathbf{A}}
\newcommand{\Bb}{\mathbf{B}}
\newcommand{\Eb}{\mathbf{E}}
\newcommand{\Ib}{\mathbf{I}}
\newcommand{\Mb}{\mathbf{M}}
\newcommand{\Sbb}{\mathbf{S}}
\newcommand{\Tb}{\mathbf{T}}
\newcommand{\Wb}{\mathbf{W}}

\newcommand{\bS}{\mathbf{S}}
\newcommand{\bU}{\mathbf{U}}
\newcommand{\bX}{\mathbf{X}}
\newcommand{\bY}{\mathbf{Y}}
\newcommand{\bZ}{\mathbf{Z}}

\newcommand{\cF}{\mathcal{F}}
\newcommand{\cN}{\mathcal{N}}

\newcommand{\bmu}{\bolds{\mu}}

\newcommand{\bDelta}{\bolds{\Delta}}
\newcommand{\bTheta}{\bolds{\Theta}}
\newcommand{\bSigma}{\bolds{\Sigma}}
\newcommand{\bOmega}{\bolds{\Omega}}

\renewcommand{\sign}{\mathop{\operatorname{sign}}}

\newcommand{\zero}{\mathbf{0}}

\newcommand{\diag}{\operatorname{diag}}

\newtheorem{theorem}{Theorem}[section]
\newtheorem{lemma}[theorem]{Lemma}
\newtheorem{proposition}[theorem]{Proposition}
\newremark{remark}[theorem]{Remark}
\newtheorem{corollary}[theorem]{Corollary}
\newproclaim{definition}[theorem]{Definition}
\makeatother

\begin{document}
\begin{frontmatter}

\title{Statistical analysis of latent generalized correlation matrix
estimation in transelliptical distribution}
\runtitle{Latent generalized correlation matrix estimation}

\begin{aug}
\author[A]{\inits{F.}\fnms{Fang}~\snm{Han}\thanksref{A}\ead[label=e1]{fhan@jhu.edu}}
\and
\author[B]{\inits{H.}\fnms{Han}~\snm{Liu}\corref{}\thanksref{B}\ead[label=e2]{hanliu@princeton.edu}}
\address[A]{Department of Biostatistics,
Johns Hopkins University,
Baltimore, MD 21205, USA.
\printead{e1}}

\address[B]{Department of Operations Research and Financial Engineering,
Princeton University,
Princeton, NJ 08544, USA.
\printead{e2}}
\end{aug}

%
\received{\smonth{11} \syear{2013}}
%
\revised{\smonth{11} \syear{2014}}

%
\begin{abstract}
Correlation matrices play a key role in many multivariate methods
(e.g., graphical model estimation and factor analysis). The current
state-of-the-art in estimating large correlation matrices focuses on
the use of Pearson's sample correlation matrix. Although Pearson's
sample correlation matrix enjoys various good properties under Gaussian
models, it is not an effective estimator when facing heavy-tailed
distributions. As a robust alternative, Han and Liu
[\textit{J.~Am. Stat. Assoc.} \textbf{109} (2015) 275--287]
advocated the use of a transformed version of the Kendall's tau sample
correlation matrix in estimating high dimensional latent generalized
correlation matrix under the transelliptical distribution family (or
elliptical copula). The transelliptical family assumes that after
unspecified marginal monotone transformations, the data follow an
elliptical distribution. In this paper, we study the theoretical
properties of the Kendall's tau sample correlation matrix and its
transformed version proposed in Han and Liu
[\textit{J.~Am. Stat. Assoc.} \textbf{109} (2015) 275--287] for
estimating the population Kendall's tau correlation matrix and the
latent Pearson's correlation matrix under both spectral and restricted
spectral norms. With regard to the spectral norm, we highlight the role
of ``effective rank'' in quantifying the rate of convergence. With
regard to the restricted spectral norm, we for the first time present a
``sign sub-Gaussian condition'' which is sufficient to guarantee that
the rank-based correlation matrix estimator attains the fast rate of
convergence. In both cases, we do not need any moment condition.
\end{abstract}

%
\end{frontmatter}

\section{Introduction}\label{sec1}

Covariance and correlation matrices play a central role in multivariate
analysis. An efficient estimation of covariance/correlation matrix is a
major step in conducting many methods, including principal component
analysis (PCA), scale-invariant PCA, graphical model estimation,
discriminant analysis, and factor analysis. Large
covariance/correlation matrix estimation receives a lot of attention in
high dimensional statistics. This is partially because the sample
covariance/correlation matrix is an inconsistent estimator when
$d/n\nrightarrow0$ ($d$ and $n$ represent the dimensionality
and sample size).

Given $n$ observations $\mathbf{x}_1,\ldots,\mathbf{x}_n$ of a
$d$-dimensional random
vector $\bX\in\reals^d$ with the population covariance matrix
$\bOmega
$, let $\hat\Sbb$ be the Pearson's sample covariance matrix calculated
based on $\mathbf{x}_1,\ldots,\mathbf{x}_n$. For theoretical
analysis, we
adopt a
similar double asymptotic framework as in Bickel and Levina \cite
{bickel2008regularized},
where we write $d$ to be the abbreviation of $d_n$, which changes with
$n$. Under this double asymptotic framework, where both the dimension
$d$ and sample size $n$ can increase to infinity, Johnstone \cite
{johnstone2001distribution}, Baik and Silverstein \cite
{baik2006eigenvalues} and Jung and Marron \cite
{jung2009pca} pointed out settings such that, even when $\bX$ follows a
Gaussian distribution with identity covariance matrix, $\hat\Sbb$ is an
inconsistent estimator of $\bSigma$ under spectral norm. In other
words, letting $\|\cdot\|_2$ denote the spectral norm of a matrix,
typically for $(n,d)\rightarrow\infty$, we have
\[
\|\hat\Sbb-\bOmega\|_2 \nrightarrow0.
\]
This observation motivates different versions of sparse
covariance/correlation matrix estimation methods. See, for example,
banding method (Bickel and Levina \cite{bickel2008regularized}),
tapering method (Cai et al. \cite
{cai2010optimal}, Cai and Zhou \cite{cai2012minimax}), and
thresholding method (Bickel and Levina \cite
{bickel2008covariance}).
However, although the regularization methods exploited are different,
they all use the Pearson's sample covariance/correlation matrix as a
pilot estimator, and accordingly the performance of the estimators
relies on existence of higher order moments of the data. For example,
letting $\|\cdot\|_{\max}$ and $\|\cdot\|_{2,s}$ denote the
element-wise supremum norm and restricted spectral norm (detailed
definitions provided later), in proving
%
\begin{equation}
\label{eq:intro2} \|\hat\Sbb-\bOmega\|_{\max}=\mathrm{O}_P \biggl(
\sqrt{\frac{\log
d}{n}} \biggr) \quad\mbox{or}\quad \|\hat\Sbb-\bOmega
\|_{2,s}=\mathrm{O}_P \biggl( \sqrt {\frac{s\log(d/s)}{n}} \biggr)
\end{equation}
(here, $d$ and $s$ are the abbreviation of $d_n$ and $s_n$ and
$\mathrm{O}_P(\cdot)$ is defined to represent the stochastic order with regard
to $n$), it is commonly assumed that, for $d=1,2,\ldots,\bX
=(X_1,\ldots,X_d)^T$ satisfies the following sub-Gaussian condition:
%
\begin{eqnarray}
\label{eq:subg-intro} \mbox{(marginal sub-Gaussian)}\quad \E\exp(tX_j) &\leq&\exp
\biggl(\frac
{\sigma^2t^2}{2} \biggr)\qquad \mbox{for all } j\in\{1,\ldots,d\}\quad\mbox{or}
\nonumber
\\[-8pt]
\\[-8pt]
\nonumber
\mbox{(multivariate sub-Gaussian)}\quad \E\exp\bigl(t\bv^T\bX\bigr)
&\leq& \exp \biggl(\frac{\sigma^2t^2}{2} \biggr)\qquad \mbox{for all } \bv\in
\S^{d-1},
\end{eqnarray}
for some absolute constant $\sigma^2>0$. Here, $\S^{d-1}$ is the
$d$-dimensional unit sphere in $\reals^d$.

The moment conditions in \eqref{eq:subg-intro} are not satisfied for
many distributions. To elaborate how strong this condition is, we
consider the student's $t$ distribution. Assuming that $T$ follows a
student's $t$ distribution with degree of freedom $\nu$, it is known
(Hogg and Craig \cite{hogg1994introduction}) that
\[
\E T^{2k} =\infty\qquad\mbox{for $k\geq\nu/2$}.
\]

Recently, Han and Liu \cite{han2012transelliptical} advocated to use the
transelliptical distribution for modeling and analyzing complex and
noisy data. They exploited a transformed version of the Kendall's tau
sample correlation matrix $\hat\bSigma$ to estimate the latent
Pearson's correlation matrix $\bSigma$. The transelliptical family
assumes that, after a set of unknown marginal transformations, the data
follow an elliptical distribution. This family is closely related to
the elliptical copula and contains many well-known distributions,
including multivariate Gaussian, rank-deficient Gaussian,
\mbox{multivariate-$t$}, Cauchy, Kotz, logistic, etc. Under the transelliptical
distribution, without any moment constraint, they showed that a
transformed Kendall's tau sample correlation matrix $\hat\bSigma$
approximates the latent Pearson's correlation matrix $\bSigma$ in a
parametric rate:
%
\begin{equation}
\label{eq:intro1} \|\hat{\bSigma}-\bSigma\|_{\max}=\mathrm{O}_P \biggl(
\sqrt{\frac{\log d}{n}} \biggr),
\end{equation}
which attains the minimax rate of convergence.

Although \eqref{eq:intro1} is inspiring, in terms of theoretical
analysis of many multivariate methods, the rates of convergence under
spectral norm and restricted spectral norm are more desired. For
example, Bickel and Levina \cite{bickel2008covariance} and Yuan and
Zhang \cite{yuan2011truncated}
showed that the performances of principal component analysis and a
computationally tractable sparse PCA method are determined by the rates
of convergence for the plug-in matrix estimators under spectral and
restricted spectral norms. A trivial extension of \eqref{eq:intro1}
gives us that
\[
\|\hat\bSigma-\bSigma\|_2=\mathrm{O}_P \biggl( d\sqrt{
\frac{\log d}{n}} \biggr) \quad\mbox{and}\quad \|\hat\bSigma-\bSigma\|_{2,s}=\mathrm{O}_P
\biggl( s\sqrt {\frac{\log d}{n}} \biggr),
\]
which are both not tight compared to the parametric rates (for more
details, check Lounici \cite{lounici2012high} and Bunea and Xiao \cite
{bunea2012sample} for
results under the spectral norm, and Vu and Lei \cite{vu2012minimax}
for results
under the restricted spectral norm).

In this paper, we push the results in Han and Liu \cite{han2012transelliptical}
forward, providing improved results of the transformed Kendall's tau
correlation matrix under both spectral and restricted spectral norms.
We consider the statistical properties of the Kendall's tau sample
correlation matrix $\hat\Tb$ in estimating the Kendall's tau
correlation matrix $\Tb$, and the transformed version $\hat\bSigma$ in
estimating~$\bSigma$.

First, we considering estimating the Kendall's tau correlation matrix
$\Tb$ itself. Estimating Kendall's tau is of its self-interest. For
example, Embrechts et al. \cite{embrechts2003modelling} claimed that
in many cases in
modeling dependence Pearson's correlation coefficient ``might prove
very misleading'' and advocated to use the Kendall's tau correlation
coefficient as the ``perhaps best alternatives to the linear
correlation coefficient as a measure of dependence for nonelliptical
distributions.'' In estimating $\Tb$, we show that, without any
condition, for any continuous random vector~$\bX$,
\[
\|\hat\Tb-\Tb\|_2=\mathrm{O}_P \biggl(\|\Tb\|_2
\sqrt{\frac{r_e(\Tb
)\log
d}{n}} \biggr),
\]
where $r_e(\Tb):=\tr(\Tb)/\|\Tb\|_2$ is called effective rank.
Moreover, we provide a new term called ``sign sub-Gaussian condition,''
under which we have
\[
\|\hat\Tb-\Tb\|_{2,s}=\mathrm{O}_P \biggl( \|\Tb\|_2
\sqrt{\frac
{s\log
d}{n}} \biggr).
\]

Secondly, under the transelliptical family, we consider estimating the
Pearson's correlation matrix $\bSigma$ of the latent elliptical
distribution using the transformed Kendall's tau sample correlation
matrix $\hat\bSigma=[\sin(\frac{\pi}{2}\hat\Tb_{jk})]$. Without any
moment condition, we show that, as long as $\bX$ belongs to the
transelliptical family,
\[
\|\hat\bSigma-\bSigma\|_2=\mathrm{O}_P \biggl(\|\bSigma
\|_2 \biggl\{ \sqrt{\frac
{r_e(\bSigma)\log d}{n}}+\frac{r_e(\bSigma)\log d}{n} \biggr\}
\biggr),
\]
which attains the nearly optimal rate of convergence obtained in
Lounici \cite
{lounici2012high} and Bunea and Xiao \cite{bunea2012sample}. Moreover,
provided that
the sign sub-Gaussian condition is satisfied, we have
\[
\|\hat\bSigma-\bSigma\|_{2,s}=\mathrm{O}_P \biggl(\|\bSigma
\|_2\sqrt {\frac
{s\log d}{n}}+\frac{s\log d}{n} \biggr),
\]
which attains the nearly optimal rate of convergence obtained in Vu and
Lei \cite
{vu2012minimax}.

\subsection{Discussion with related works}\label{sec1.1}

Our work is related to a vast literature in large covariance matrix
estimation, with different settings of sparsity assumptions (Cai et al.
\cite
{cai2010optimal,cai2013optimal}, Cai and Zhou \cite{cai2012minimax}, Vu and Lei \cite
{vu2012minimax}), or
without any sparsity assumption (Bunea and Xiao \cite{bunea2012sample},
Lounici \cite{lounici2012high}). In particular, this work is closely
related to Lounici \cite{lounici2012high} and Bunea and Xiao \cite
{bunea2012sample} with
regard to the theoretical analysis of the spectral norm convergence,
and the work of Vu and Lei \cite{vu2012minimax} with regard to the theoretical
analysis of the restricted spectral norm convergence.

However, there are various new contributions made in this paper given
the aforementioned results. We emphasize the advantage of rank-based
statistics over moment-based statistics. One new message delivered in
this paper is, via resorting to the rank-based statistics, the
statistical efficiency attained by the aforementioned methods under
some stringent moment constraints, can be attained under some more
flexible models. Moreover, we believe that the technical developments
built in this paper, including the analysis of $U$-statistics, the
concentration of matrix-value functions, and the verification of the
sign sub-Gaussian condition for several particular models, are distinct
from the existing literature and of self-interest.

Our work is also closely related to an expanding literature in
extending copula models to the high dimensional settings. These include
the use of the nonparanormal (Gaussian copula) and the transelliptical
(elliptical copula) distribution families. Methodologically, the
Spearman's rho is recommended in the analysis of the nonparanormal
family for conducting graphical model estimation (Liu et al. \cite
{liu2012high}, Xue and Zou \cite{xue2012regularized}), classification
(Han et al. \cite{han2013coda}),
and PCA (Han and Liu \cite{han2012semiparametric}). The Kendall's tau
is recommended
in the analysis of the transelliptical family for conducting graphical
model estimation (Liu et al. \cite{liu2012transelliptical}) and PCA
(Han and Liu \cite
{han2012transelliptical}).

Our work is motivated from the aforementioned results. But, different
from the existing ones, we give a more general study on the convergence
of the Kendall's tau matrix itself, and provide more insights into the
rank-based statistics. We characterize three types of convergence with
regard to the Kendal's tau matrix $\hat\Tb$ and its transformed version
$\hat\bSigma$: The element-wise supremum norm ($\ell_{\max}$), the
spectral norm ($\ell_{2}$), and the restricted spectral norm ($\ell
_{2,s}$). In comparison, the existing results only exploited the $\ell
_{\max}$ convergence result, which we find is not sufficient in showing
the statistical efficiency of many rank-based methods. It is also worth
noting that the new theories developed here with regard to the $\ell_2$
and $\ell_{2,s}$ convergence have broad implications. They can be
easily applied to the study of factor model, sparse PCA, robust
regression and many other methods, and can lead to more refined
statistical analysis.

In an independent work, Wegkamp and Zhao \cite{marten2013} proposed to
use the same
transformed Kendall's tau correlation coefficient estimator to analyze
the elliptical copula factor model and proved a similar spectral norm
convergence result as in Theorem~\ref{thm:lowd_spectral2} of this paper. The proofs are
different and these two papers are independent work. 

\subsection{Notation system}\label{sec1.2}

Let $\Mb=[\Mb_{ij}] \in\mathbb{R}^{d \times d}$ and $\bv
=(v_1,\ldots,v_d)^{T} \in\mathbb{R}^d$. We denote $\bv_I$ to be the
subvector of $\bv$ whose entries are indexed by a set $I$. We also
denote $\Mb_{I,J}$ to be the submatrix of $\Mb$ whose rows are indexed
by $I$ and columns are indexed by $J$. Let $\Mb_{I*}$ and $\Mb_{*J}$ be
the submatrix of $\Mb$ with rows indexed by $I$, and the submatrix of
$\Mb$ with columns indexed by $J$. Let $\supp(\bv):=\{j\dvt v_j\ne
0\}$. For
$0 < q < \infty$, we define the $\ell_0$, $\ell_q$, and $\ell
_{\infty}$
vector (pseudo-)norms as
\[
\|\bv\|_0:=\card\bigl(\supp(\bv)\bigr), \qquad\|\bv\|_q:=
\Biggl(\sum_{i=1}^d |v_i|^q
\Biggr)^{1/q} \quad\mbox{and} \quad\|\bv\|_{\infty}:=\max
_{1\leq i
\leq d} |v_i|.
\]
Let $\lambda_j(\Mb)$ be the $j$th largest eigenvalue of $\Mb$ and
$\bTheta_j(\Mb)$ be a corresponding eigenvector. In particular, we let
$\lambda_{\max}(\Mb):=\lambda_1(\Mb)$. We define $\S^{d-1}:= \{
\bv\in
\reals^d\dvt\|\bv\|_2=1 \}$ to be the $d$-dimensional unit
sphere. We
define the matrix element-wise supremum norm ($\ell_{\max}$ norm),
spectral norm ($\ell_2$ norm), and restricted spectral norm ($\ell
_{2,s}$ norm) as
\[
\|\Mb\|_{\max}:=\max\bigl\{|\Mb_{ij}|\bigr\},\qquad \|\Mb\|_2:=
\sup_{\bv
\in\S
^{d-1}}\|\Mb\bv\|_2\quad \mbox{and} \quad\|\Mb
\|_{2,s}:=\sup_{\bv\in\S
^{d-1}\cap\|\bv\|_0\leq s}\|\Mb\bv\|_2.
\]
We define $\diag(\Mb)$ to be a diagonal matrix with $[\diag(\Mb
)]_{jj}=\Mb_{jj}$ for $j=1,\ldots,d$. We also denote $\vec(\Mb
):=(\Mb
_{* 1}^T, \ldots, \Mb_{* d}^T)^T$. For any two vectors $\ba,\mathbf
{b}\in\reals
^d$, we denote $\langle\ba,\mathbf{b} \rangle:=\ba^T\mathbf{b}$
and $\sign(\ba
):=(\sign
(a_1),\ldots,\sign(a_d))^T$, where $\sign(x)=x/|x|$ with the
convention $0/0=0$.

\subsection{Paper organization}\label{sec1.3}

The rest of this paper is organized as follows. In the next section, we
briefly overview the transelliptical distribution family and the main
concentration results for the transformed Kendall's tau sample
correlation matrix proposed by Han and Liu \cite
{han2012transelliptical}. In
Section~\ref{sec:theory1}, we analyze the convergence rates of Kendall's tau sample
correlation matrix and its transformed version with regard to the
spectral norm. In Section~\ref{sec:theory2}, we analyze the convergence rates of
Kendall's tau sample correlation matrix and its transformed version
with regard to the restricted spectral norm. The technical proofs of
these results are provided in Section~\ref{sec:proof}. More discussions and
conclusions are provided in Section~\ref{sec6}.

\section{Preliminaries and background overview}\label{sec2}

In this section, we briefly review the transelliptical distribution and
the corresponding latent generalized correlation matrix estimator
proposed by Han and Liu \cite{han2012transelliptical}.

\subsection{Transelliptical distribution family}\label{sec2.1}

The concept of transelliptical distribution builds upon the elliptical
distribution. Accordingly, we first provide a definition of the
elliptical distribution, using the stochastic representation as in Fang
et al. \cite
{fangsymmetric}. In the sequel, for any two random vectors $\bX$ and
$\bY$, we denote $\bX\stackrel{d}{=}\bY$ if they are identically
distributed.
%
\begin{definition}[(Fang et al. \cite{fangsymmetric})]\label{def:ec1}
A random vector $\bZ=(Z_1,\ldots,Z_d)^T$ follows an elliptical
distribution if and only if $\bZ$ has a stochastic representation:
$\bZ\stackrel{d}{=}\bmu+\xi\Ab\bU$. Here $\bmu\in\reals^d$,
$q:=\rank(\Ab)$, $\Ab\in\reals^{d\times q}$, $\xi\geq0$ is a
random variable independent of $\bU$, $\bU\in\S^{q-1}$ is uniformly
distributed on the unit sphere in $\reals^q$. In this setting, letting
$\bSigma:=\Ab\Ab^T$, we denote $\bZ\sim \mathit{EC}_d(\bmu,\bSigma,\xi)$. Here,
$\bSigma$ is called the scatter matrix.
\end{definition}

The elliptical family can be viewed as a semiparametric generalization
of the Gaussian family, maintaining the symmetric property of the
Gaussian distribution but allowing heavy tails and richer structures.
Moreover, it is a natural model for many multivariate methods such as
principal component analysis (Boente et al. \cite
{boente2012characterization}). The
transelliptical distribution family further relaxes the symmetric
assumption of the elliptical distribution by assuming that, after
unspecified strictly increasing marginal transformations, the data are
elliptically distributed. A formal definition of the transelliptical
distribution is as follows.

\begin{definition}[(Han and Liu \cite{han2012transelliptical})]\label{def:TE}
A random vector $\bX=(X_1,\ldots,X_d)^T$ follows a
transelliptical distribution, denoted by $\bX\sim
\mathit{TE}_d(\bSigma,\xi; f_1,\ldots,f_d)$, if there exist univariate
strictly increasing functions $f_{1},\ldots, f_{d}$ such that
\begin{eqnarray*}
\bigl(f_1(X_1),\ldots,f_d(X_d)
\bigr)^T \sim \mathit{EC}_d(\zero,\bSigma,\xi)\qquad \mbox{where }
\diag(\bSigma)=\Ib _d \mbox { and } \P(\xi=0)=0.
\end{eqnarray*}
Here $\Ib_d\in\reals^{d\times d}$ is the $d$-dimensional identity
matrix and $\bSigma$ is called the {\it latent generalized correlation
matrix}.
\end{definition}

We note that the transelliptical distribution is closely related to the
nonparanormal distribution (Liu et al. \cite
{liu2009nonparanormal,liu2012high}, Xue and Zou
\cite{xue2012regularized}, Han and Liu \cite{han2012semiparametric},
Han et al. \cite{han2013coda})
and meta-elliptical distribution (Fang et al. \cite{fang2002}). The
nonparanormal
distribution assumes that after unspecified strictly increasing
marginal transformations the data are Gaussian distributed. It is easy
to see that the transelliptical family contains the nonparanormal
family. On the other hand, it is subtle to elaborate the difference
between the transelliptical and meta-elliptical. In short, the
transelliptical family contains meta-elliptical family. Compared to the
meta-elliptical, the transelliptical family does not require the random
vectors to have densities and brings new insight into both theoretical
analysis and model interpretability. We refer to Liu et al. \cite
{liu2012transelliptical} for more detailed discussion on the comparison
between the transelliptical family, nonparanormal and meta-elliptical families.

\subsection{Latent generalized correlation matrix estimation}\label{sec2.2}

Following Han and Liu \cite{han2012transelliptical}, we are interested in
estimating the latent generalized correlation matrix $\bSigma$, i.e.,
the correlation matrix of the latent elliptically distributed random
vector $f(\bX):=(f_1(X_1),\ldots,f_d(X_d))^T$.
By treating both the generating variable $\xi$ and the marginal
transformation functions $f=\{f_j\}_{j=1}^d$ as nuisance parameters,
Han and Liu \cite{han2012transelliptical} proposed to use a
transformed Kendall's
tau sample correlation matrix to estimate the latent generalized
correlation matrix $\bSigma$. More specifically, letting $\mathbf
{x}_1,\ldots
,\mathbf{x}_n$ be $n$ independent and identically distributed
observations of
a random vector $\bX\in \mathit{TE}_d(\bSigma,\xi;f_1,\ldots,f_d)$, the
Kendall's tau correlation coefficient between the variables $X_{j}$ and
$X_{k}$ is defined as
\[
\hat{\tau}_{jk}:=\frac{2}{n(n-1)}\sum_{i<i'}
\sign \bigl((\mathbf {x}_{i}-\mathbf{x} _{i'})_{j}(
\mathbf{x}_{i}-\mathbf{x}_{i'})_{k} \bigr).
\]
Its population quantity can be written as
%
\begin{equation}
\label{eq:kendall} \tau_{jk}:=\P \bigl((X_j-
\tilde{X}_j) (X_k-\tilde{X}_k)>0\bigr)-\P
\bigl((X_j-\tilde {X}_j) (X_k-
\tilde{X}_k)<0 \bigr),
\end{equation}
where $\tilde{\bX}=(\tilde{X}_1,\ldots,\tilde{X}_d)^T$ is an
independent copy of $\bX$. We denote
\[
\Tb:=[\tau_{jk}] \quad\mbox{and}\quad \hat\Tb:=[\hat\tau_{jk}]
\]
to be the Kendall's tau correlation matrix and Kendall's tau sample
correlation matrix.

For the transelliptical family, it is known that
$\bSigma_{jk}=\sin(\frac{\pi}{2}\tau_{jk})$ (check, e.g., Theorem~3.2
in Han and Liu \cite{han2012transelliptical}). A latent generalized correlation
matrix estimator $\hat{\bSigma}:=[\hat{\bSigma}_{jk}]$, called the
transformed Kendall's tau sample correlation matrix, is accordingly
defined by
%
\begin{equation}
\label{eq:hatsigma} \hat{\bSigma}_{jk}=\sin \biggl(\frac{\pi}{2}\hat{
\tau}_{jk} \biggr).
\end{equation}
%
Han and Liu \cite{han2012transelliptical} showed that, without any
moment constraint,
\[
\|\hat\bSigma-\bSigma\|_{\max}=\mathrm{O}_P \biggl( \sqrt{
\frac{\log d}{n}} \biggr),
\]
and accordingly by simple algebra we have
%
\begin{equation}
\label{eq:sec2-1} \|\hat\bSigma-\bSigma\|_{2}=\mathrm{O}_P \biggl( d
\sqrt{\frac{\log
d}{n}} \biggr) \quad\mbox{and}\quad \|\hat\bSigma-\bSigma
\|_{2,s}=\mathrm{O}_P \biggl( s\sqrt{\frac{\log d}{n}} \biggr).
\end{equation}
The rates of convergence in \eqref{eq:sec2-1} are far from optimal
(check Lounici \cite{lounici2012high}, Bunea and Xiao \cite
{bunea2012sample}, and Vu and Lei \cite
{vu2012minimax} for the parametric rates). In the next two sections, we
will push the results in Han and Liu \cite{han2012transelliptical}
forward, showing
that better rates of convergence can be built in estimating the
Kendall's tau correlation matrix and the latent generalized correlation matrix.

\section{Rate of convergence under spectral norm}
\label{sec:theory1}

In this section, we provide the rate of convergence of the Kendall's
tau sample correlation matrix $\hat\Tb$ to $\Tb$, as well as the
transformed Kendall's tau sample correlation matrix $\hat\bSigma$ to
$\bSigma$, under the spectral norm. The next theorem shows that,
without any moment constraint or assumption on the data distribution
(as long as it is continuous), the rate of convergence of $\hat{\Tb}$
to $\Tb$ under the spectral norm is $\|\Tb\|_2\sqrt{r_e(\Tb
)\log
d/n}$, where for any positive semidefinite matrix $\Mb\in\reals
^{d\times d}$,
\[
r_e(\Mb):=\frac{\tr(\Mb)}{\|\Mb\|_2}
\]
is called the effective rank of $\Mb$ and must be less than or equal to
the dimension $d$. For notational simplicity, in the sequel we assume
that the sample size $n$ is even. When $n$ is odd, we can always use
$n-1$ data points without affecting the obtained rate of convergence.

\begin{theorem}\label{thm:lowd_spectral2} Let $\mathbf{x}_1,\ldots
,\mathbf
{x}_n$ be
$n$ observations of a $d$-dimensional continuous random vector~$\bX$.
Then when $r_e(\Tb)\log d/n\rightarrow0$, for sufficiently large $n$
and any $0<\alpha<1$, with probability larger than $1-2\alpha$, we have
%
\begin{equation}
\label{eq:TB} \|\hat{\Tb}-\Tb\|_2 \leq4\|\Tb\|_2\sqrt{
\frac{\{r_e(\Tb
)+1\}\log
(d/\alpha)}{3n}}.
\end{equation}
\end{theorem}

Theorem~\ref{thm:lowd_spectral2} shows that, when $r_e(\Tb)\log d/n
\rightarrow0$, we have
\[
\|\hat{\Tb}-\Tb\|_{2}=\mathrm{O}_P \biggl(\|\Tb\|_2
\sqrt{\frac
{r_e(\Tb)\log
d}{n}} \biggr).
\]
This rate of convergence we proved is the same parametric rate as
obtained in Vershynin \cite{vershynin2010introduction}, Lounici \cite
{lounici2012high},
and Bunea and Xiao \cite{bunea2012sample} when there is not any
additional structure.%

In the next theorem, we show that, under the modeling assumption that
$\bX$ is transelliptically distributed, which is of particular interest
in real applications as shown in Han and Liu \cite
{han2012transelliptical}, we have
that a transformed version of the Kendall's tau sample correlation
matrix can estimate the latent generalized correlation matrix in a
nearly optimal rate.

\begin{theorem}\label{thm:lowd_spectral} Let $\mathbf{x}_1,\ldots
,\mathbf
{x}_n$ be
$n$ observations of $\bX\sim \mathit{TE}_d(\bSigma,\xi;f_1,\ldots,f_d)$. Let
$\hat{\bSigma}$ be the transformed Kendall's tau sample correlation
matrix defined in \eqref{eq:hatsigma}. We have, when $r_e(\bSigma
)\log
d/n\rightarrow0$, for $n$ large enough and $0<\alpha<1$, with
probability larger than $1-2\alpha-\alpha^2$,
%
\begin{equation}
\label{eq:theoremlowd} \|\hat{\bSigma}-\bSigma\|_2 \leq\pi^2\|
\bSigma\|_2 \biggl(2\sqrt {\frac{\{r_e(\bSigma)+1\}\log(d/\alpha)}{3n}}+\frac{r_e(\bSigma
)\log
(d/\alpha)}{n}
\biggr).
\end{equation}
%
\end{theorem}

Theorem~\ref{thm:lowd_spectral} indicates that, when $r_e(\bSigma
)\log
d/n \rightarrow0$, we have
\[
\|\hat{\bSigma}-\bSigma\|_{2}=\mathrm{O}_P \biggl(\|\bSigma
\|_2\sqrt {\frac
{r_e(\bSigma)\log d}{n}} \biggr).
\]
By the discussion of Theorem~2 in Lounici \cite{lounici2012high}, the obtained
rate of convergence is minimax optimal up to a logarithmic factor with
respect to a suitable parameter space. However, compared to the
conditions in Lounici \cite{lounici2012high}, and Bunea and Xiao \cite
{bunea2012sample}, which
require strong multivariate sub-Gaussian modeling assumption on $\bX$
(which implies the existence of moments of arbitrary order), $\hat
\bSigma$ attains this parametric rate in estimating the latent
generalized correlation matrix without any moment constraints.

\begin{remark} The $\log d$ term presented in the rate of convergence
of $\hat\Tb$ and $\hat\bSigma$ is an artifact of the proof, and also
appears in the statistical analysis of the sample covariance matrix
under the sub-Gaussian model (see, e.g., Proposition~3 in Lounici \cite
{lounici2012high} and Theorem~2.2 in Bunea and Xiao \cite
{bunea2012sample}). If we
would like to highlight the role of the effective rank, $r_e(\Tb)$ and
$r_e(\bSigma)$, to our knowledge there is no work that can avoid the
$\log d$ term. On the other hand, in estimating $\Tb$ using $\hat\Tb$,
a $\mathrm{O}_P(\sqrt{d/n})$ rate of convergence can be attained under the
condition of Theorem~\ref{thm:main} provided in the next section.
In estimating $\bSigma$ using $\hat\bSigma$, a $\mathrm{O}_P(\sqrt{d/n})$ rate
of convergence is also attainable under the condition of Theorem~\ref
{thm:main} when $d(\log d)^2=\mathrm{O}(n)$.
\end{remark}

\section{Rate of convergence under restricted spectral norm}
\label{sec:theory2}

In this section, we analyze the rates of convergence of the Kendall's
tau sample correlation matrix and its transformed version under the
restricted spectral norm. The main target is to improve the rate
$\mathrm{O}_P(s\sqrt{\log d/n})$ shown in \eqref{eq:sec2-1} to the rate
$\mathrm{O}_P(\sqrt{s\log(d/s)/n})$. Such a rate has been shown to be minimax
optimal under the Gaussian model (via combining Theorem~2.1 and Lemma~3.2.1 in Vu and Lei \cite{vu2012minimax}). Obtaining such an improved
rate is
technically challenging since the data could be very heavy-tailed and
the transformed Kendall's tau sample correlation matrix has a much more
complex structure than the Pearson's covariance/correlation matrix.




In the following, we lay out a venue to analyze the statistical
efficiency of $\hat\Tb$ and $\hat\bSigma$ under the restricted spectral
norm. In particular, we characterize a subset of the transelliptical
distributions for which $\hat\Tb$ and $\hat\bSigma$ can approximate
$\Tb
$ and $\bSigma$ in an improved rate. More specifically, we provide a
``sign sub-Gaussian'' condition which is sufficient for $\hat{\Tb}$ and
$\hat{\bSigma}$ to attain the nearly optimal rate. This condition is
related to the sub-Gaussian assumption in Vu and Lei \cite
{vu2012minimax}, Lounici \cite
{lounici2012high}, and Bunea and Xiao \cite{bunea2012sample} (see
Assumption~2.2 in
Vu and Lei \cite{vu2012minimax}, e.g.). Before proceeding to
the formal
definition of this condition, we first define an operator $\psi\dvtx
\reals
\rightarrow\reals$ as follows.
%
\begin{definition}\label{def:orlicz} For any random variable $Y\in
\reals
$, the operator $\psi\dvtx\reals\rightarrow\reals$ is defined as
%
\begin{equation}
\label{eq:subgaussian0} \psi(Y;\alpha,t_0):= \inf \bigl\{ c>0\dvt\E\exp \bigl
\{ t\bigl(Y^\alpha-\E Y^\alpha\bigr) \bigr\}\leq\exp
\bigl(ct^2\bigr), \mbox{for } |t|<t_0 \bigr\}.
\end{equation}
\end{definition}
The operator $\psi(\cdot)$ can be used to quantify the tail behaviors
of random variables. We recall that a zero-mean random variable $X\in
\reals$ is said to be sub-Gaussian if there exists a constant $c$ such
that $\E\exp(tX)\leq\exp(ct^2)$ for all $t\in\reals$. A zero-mean
random variable $Y\in\reals$ with $\psi(Y;1,\infty)$ bounded is well
known to be sub-Gaussian, which implies a tail probability
\[
\P\bigl(|Y-\E Y|>t\bigr)<2\exp\bigl(-t^2/(4c)\bigr),
\]
where $c$ is the constant defined in equation \eqref{eq:subgaussian0}.
Moreover, $\psi(Y;\alpha,t_0)$ is related to the
Orlicz $\psi_{2}$-norm. A formal definition of the Orlicz norm is
provided as follows.
%
\begin{definition}\label{def:orliczz}
For any random variable $Y\in
\reals
$, its Orlicz $\psi_{2}$-norm is defined as
\[
\|Y\|_{\psi_{2}} := \inf \bigl\{ c>0\dvt\E\exp \bigl( |Y/c|^2
\bigr)\leq2 \bigr\}.
\]
\end{definition}
It is well known that a random variable $Y$ has $\psi(Y;1,\infty)$ to
be bounded if and only if $\|Y\|_{\psi_2}$ in Definition~\ref{def:orliczz} is bounded (van de Geer and Lederer \cite
{van2011bernstein}). We refer to Lemma~\ref
{lem:subgaussian} in the \hyperref[app]{Appendix} for a more detailed
description on
this property.

Another relevant norm to $\psi(\cdot)$ is the sub-Gaussian norm
$\|\cdot\|_{\phi_2}$ used in, for example, Vershynin~\cite
{vershynin2010introduction}. A former definition of the sub-Gaussian
norm is as follows.
%
\begin{definition}\label{def:subgaussian} For any random variable
$X\in
\reals$, its sub-Gaussian norm is defined as
\[
\|X\|_{\phi_{2}} := \sup_{k\geq1}k^{-1/2} \bigl(
\E|X|^k \bigr)^{1/k}.
\]
\end{definition}
The sub-Gaussian norm is also highly related to the sub-Gaussian random
variables. In particular, we have if $\E X=0$, then $\E\exp(tX)\leq
\exp
(Ct^2\|X\|^2_{\phi_2})$.



Using the operator $\psi(\cdot)$, we now proceed to define the sign
sub-Gaussian condition. For mathematical rigorousness, the formal
definition is posed on $\{\cF^d, d=1,2,\ldots\}$, where $\cF^d$
represents a set of probability measures on $\reals^d$. Here for any
vector $\bv=(v_1,\ldots,v_d)\in\reals^d$, we remind that $\sign
(\bv
):=(\sign(v_1),\ldots,\sign(v_d))^T$. In the following, a random vector
$\bX$ is said to be in a set of probability measures $\cF'$ if its
distribution is in $\cF'$.

\begin{definition}[(Sign sub-Gaussian condition)]\label
{def:signsubgaussian} For $d=1,2,\ldots,$ let $\cF^d$ be a set of
probability measures on $\reals^d$ such that infinitely many sets $\cF
^d$ are nonempty and $\cF:=\bigcup_{d=1}^{\infty}\cF^d$. $\cF$ is
said to
satisfy the sign sub-Gaussian condition if and only if for any $\bX$ in
$\cF$, we have
%
\begin{equation}
\label{eq:subgaussian} \sup_{\bv\in\S^{d-1}}\psi \bigl(\bigl\langle\sign(\bX-
\tilde{\bX }),\bv \bigr\rangle;2,t_0 \bigr) \leq K\|\Tb
\|_2^2,
\end{equation}
where $\tilde\bX$ is an independent copy of $\bX$, $K$ is an absolute
constant, and $t_0$ is another absolute positive number such that
$t_0\|\Tb\|_2$ is lower bounded by an absolute positive constant. We
remind that here $\Tb$ can be written as
\[
\Tb:=\E\sign(\bX-\tilde\bX)\cdot\bigl(\sign(\bX-\tilde\bX)\bigr)^T.
\]
\end{definition}

To gain more insights about the sign sub-Gaussian condition, we point
out two sets of probability measures of interest that satisfy the sign
sub-Gaussian condition.

\begin{proposition}\label{prop:example1} Suppose the set of probability
measures $\cF$ satisfies that for any random vector $\bX$ in $\cF$ and
$\tilde{\bX}$ being an independent copy of $\bX$, we have
%
\begin{equation}
\label{eq:subgaussian2} \sup_{\bv\in\S^{d-1}}\bigl\llVert \bigl\langle\sign(\bX-
\tilde{\bX }),\bv \bigr\rangle^2-\bv ^T\Tb\bv\bigr\rrVert
_{\psi_2} \leq L_1\|\Tb\|_2,
\end{equation}
where $L_1$ is a fixed constant. Then $\cF$ satisfies the sign
sub-Gaussian condition by setting $t_0=\infty$ and $K=5L_1^2/2$ in
equation \eqref{eq:subgaussian}.
\end{proposition}

\begin{proposition}\label{prop:example2} Suppose the set of probability
measure $\cF$ satisfies that for any random vector $\bX$ in $\cF$ and
$\tilde{\bX}$ being an independent copy of $\bX$, we have there exists
an absolute constant $L_2$ such that
%
\begin{equation}
\label{eq:subgaussian3} \bigl\|\bv^T\sign(\bX-\tilde{\bX})\bigr\|_{\phi_2}^2
\leq\frac
{L_2\|\Tb \|_2}{2} \qquad\mbox{for all } \bv\in\S^{d-1}.
\end{equation}
Then $\cF$ satisfies the sign sub-Gaussian condition with $t_0=c\|\Tb
\|_2^{-1}$ and $K=C$ in equation \eqref{eq:subgaussian}, where $c$
and $C$ are two fixed absolute constants.
\end{proposition}

In the following, for clarity of presentation, we abuse notation a
little and write that $\bX$ satisfies the sign sub-Gaussian condition
if there exists a set of probability measures $\cF$ satisfying the sign
sub-Gaussian condition such that for $d=1,2,\ldots,\bX\in\reals^d$
is in $\cF$.

Proposition~\ref{prop:example2} builds a bridge between the sign
sub-Gaussian condition and Assumption~1 in Bunea and Xiao \cite
{bunea2012sample} and
Lounici \cite{lounici2012high}. More specifically, saying that $\bX$ satisfies
equation \eqref{eq:subgaussian3} is equivalent to saying that $\sign
(\bX
-\tilde{\bX})$ satisfies the multivariate sub-Gaussian condition
defined in Bunea and Xiao \cite{bunea2012sample}. Therefore,
Proposition~\ref
{prop:example2} can be treated as an explanation of why we call the
condition in equation \eqref{eq:subgaussian} ``sign sub-Gaussian.''
However, by Lemma~5.14 in Vershynin \cite{vershynin2010introduction},
the sign
sub-Gaussian condition is weaker than that of equation \eqref
{eq:subgaussian3}, that is, a set of probability measures satisfying
the sign sub-Gaussian condition does not necessarily satisfy the
condition in Proposition~\ref{prop:example2}.

The sign sub-Gaussian condition is intuitive due to its relation to the
Orlicz and sub-Gaussian norms. However, it is extremely difficult to
verify whether a given set of distributions satisfies this condition.
The main difficulty lies in the fact that we must sharply characterize
the tail behavior of the summation of a sequence of possibly correlated
discrete Bernoulli random variables, which is much harder than
analyzing the summation of Gaussian random variables as usually done in
the literature.

In the following, we provide several examples of sets of distributions
that satisfy the sign sub-Gaussian condition. The next theorem shows
that the transelliptically distributed random vector $\bX\sim
\mathit{TE}_d(\bSigma,\xi; f_1,\ldots,f_d)$ such that $\bSigma=\Ib_d$
(i.e., the
underlying is a spherical distribution) for $d=1,2,\ldots$ satisfies
the sign sub-Gaussian condition. The proof of Theorem~\ref{thm:example}
is in Section~\ref{sec:example1}.

\begin{theorem}\label{thm:example} Suppose that, for $d=1,2,\ldots
,\bX\sim
\mathit{TE}_d(\Ib_d,\xi; f_1,\ldots,f_d)$ is transelliptically distributed with
a latent spherical distribution. Then $\bX$ satisfies the sign
sub-Gaussian condition.
\end{theorem}

In the next theorem, we provide a stronger version of Theorem~\ref
{thm:example}. We call a square matrix compound symmetric if the
off-diagonal values of the matrix are equal. The next theorem shows
that the transelliptically distributed $\bX\sim \mathit{TE}_d(\bSigma,\xi
;f_1,\ldots,f_d)$, with $\bSigma$ a compound symmetric matrix,
satisfies equation \eqref{eq:subgaussian3} and, therefore, satisfies
the sign sub-Gaussian condition.

\begin{theorem}\label{thm:example2} Suppose that for $d=1,2,\ldots
,\bX\sim
\mathit{TE}_d(\bSigma,\xi; f_1,\ldots,f_d)$ is transelliptically distributed
such that $\bSigma$ is a compound symmetric matrix (i.e., $\bSigma
_{jk}=\rho$ for all $j\neq k$). Then if $0 \leq\rho:=\bSigma
_{12}\leq
C_0<1$ for some absolute positive constant $C_0$, we have that $\bX$
satisfies the sign sub-Gaussian condition.
\end{theorem}

Although Theorem~\ref{thm:example} can be directly proved using the
result in Theorem~\ref{thm:example2}, the proof of Theorem~\ref
{thm:example} contains utterly different techniques which are more
transparent and illustrate the main challenges of analyzing binary
sequences even in the uncorrelated setting. Therefore, we still list
this theorem separately and provide a separate proof in Section~\ref
{sec:example1}. Theorem~\ref{thm:example2} leads to the following
corollary, which characterizes a subfamily of the transelliptical
distributions satisfying the sign sub-Gaussian condition.
%
\begin{corollary}\label{cor:compound}Suppose that for $d=1,2,\ldots
,\bX\sim
\mathit{TE}_d(\bSigma,\xi; f_1,\ldots,f_d)$ is transelliptically distributed
with $\bSigma$ a block diagonal compound symmetric matrix, that is,
%
\begin{equation}
\label{eq:compound} \bSigma= %
\pmatrix{ \bSigma_1 & \zero&
\zero& \ldots& \zero\vspace *{2pt}
\cr
\zero& \bSigma_2 & \zero&
\ldots& \zero\vspace*{2pt}
\cr
\vdots& \ddots& \cdots& \cdots& \vdots\vspace*{2pt}
\cr
\zero& \zero& \zero& \ldots& \bSigma_{q} } %
,
\end{equation}
where $\bSigma_k\in\reals^{d_k\times d_k}$ for $k=1,\ldots,q$ is
compound symmetric matrix with $\rho_k:=[\bSigma_k]_{12}\geq0$. We
have, if $q$ is upper bounded by an absolute positive constant and
$0\leq\rho_k\leq C_1<1$ for some absolute positive constant $C_1$,
$\bX
$ satisfies the sign sub-Gaussian condition.
\end{corollary}
We call the matrix in the form of equation \eqref{eq:compound} block
diagonal compound symmetric matrix. Corollary~\ref{cor:compound}
implies that transelliptically distributed random vectors with a latent
block diagonal compound symmetric latent generalized correlation matrix
satisfy the sign sub-Gaussian condition.

\begin{remark}
The sub-Gaussian condition is an artifact of the proof. Right now, we
are not aware of any transelliptical distribution that does not satisfy
this condition. More investigation on the necessity of this condition
is challenging due to the discontinuity issue of the sign
transformation and will be left for future investigation.
\end{remark}

Using the sign sub-Gaussian condition, we have the following main
result, which shows that as long as the sign sub-Gaussian condition
holds, improved rates of convergence for both $\hat\Tb$ and $\hat
\bSigma
$ under the restricted spectral norm can be attained.


\begin{theorem}\label{thm:main} For $d=1,2,\ldots,$ let $\mathbf
{x}_1,\ldots
,\mathbf{x}_n$ be $n$ observations of $\bX\in\reals^d$, for which
the sign
sub-Gaussian condition holds. We have, when $s\log(d/s)/n\rightarrow
0$, with probability larger than $1-2\alpha$,
%
\begin{equation}
\label{eq:thmmain1} \|\hat\Tb-\Tb\|_{2,s}\leq4(2K)^{1/2}\|\Tb
\|_2\sqrt{\frac
{s(3+\log
(d/s))+\log(1/\alpha)}{n}}.
\end{equation}
Moreover, when we further have $\bX\sim \mathit{TE}_d(\bSigma,\xi;f_1,\ldots
,f_d)$, with probability larger $1-2\alpha-\alpha^2$,
%
\begin{equation}
\label{eq:thmmain2} \|\hat\bSigma-\bSigma\|_{2,s}\leq\pi^2
\biggl(2(2K)^{1/2}\|\bSigma \|_2\sqrt{\frac{s(3+\log(d/s))+\log(1/\alpha)}{n}}+
\frac{s\log
(d/\alpha
)}{n} \biggr).
\end{equation}
\end{theorem}


The results presented in Theorem~\ref{thm:main} show that under various
settings the rate of convergence for $\hat\bSigma$ under the restricted
spectral norm is $\mathrm{O}_P(\sqrt{s\log(d/s)/n})$, which is the parametric
and minimax optimal rate shown in Vu and Lei \cite{vu2012minimax}
within the
Gaussian family. However, the Kendall's tau sample correlation matrix
and its transformed version attains this rate with all the moment
constraints waived.

\section{Technical proofs}\label{sec:proof}

We provide the technical proofs of the theorems shown in Sections \ref
{sec:theory1} and \ref{sec:theory2}.

\subsection{Proof of Theorem \texorpdfstring{\protect\ref{thm:lowd_spectral2}}{3.1}}\label{sec5.1}



\begin{pf}
Reminding that $\mathbf{x}_i:=(x_{i1},\ldots,x_{id})^T$, for $i\ne
i'$, let
\[
\bS_{i,i'}:=\bigl(\sign(x_{i,1}-x_{i',1}),\ldots,
\sign(x_{i,d}-x_{i',d})\bigr)^T.
\]
We denote
$\hat{\bDelta}_{i,i'}$ to be $n(n-1)$ random matrices with
\[
\hat{\bDelta}_{i,i'}:=\frac{1}{n(n-1)}\bigl(\bS_{i,i'}
\bS_{i,i'}^T-\Tb\bigr).
\]
By simple calculation, we have $\hat{\Tb}-\Tb=\sum_{i,i'}\hat
{\bDelta
}_{i,i'}$ and $\hat{\Tb}-\Tb$ is a $U$-statistic.

In the following we extend the standard decoupling trick from Hoeffding
\cite
{hoeffding1963probability} from the $U$-statistic of random variables to
the matrix setting. The extension relies on the matrix version of the
Laplace transform method. For any square matrix $\Mb\in\reals^d$, we define
\[
\exp(\Mb):=\Ib_d+\sum_{k=1}^{\infty}
\frac{\Mb^k}{k!},
\]
where $k!$ represents the factorial product of $k$. Using Proposition~3.1 in Tropp \cite{tropp2011user}, we have
%
\begin{equation}
\label{eq:LapMethodMatrix} \P\bigl[ \lambda_{\max} ( \hat{\Tb}-\Tb) \ge t\bigr] \le
\inf_{\theta> 0}\mathrm{e}^{-\theta t} \E \bigl[ \tr \mathrm{e}^{\theta(\hat{\Tb
}-\Tb
)}
\bigr],
\end{equation}
and we bound $ \E [ \tr \mathrm{e}^{\theta(\hat{\Tb}-\Tb)} ]$ as follows.

The trace exponential function
\[
\tr\exp\dvtx \Ab\to\tr \mathrm{e}^\Ab
\]
is a convex mapping from the space of self-adjoint matrix to $\mathbb
{R}^+$ (see Section~2.4 of Tropp~\cite{tropp2011user} and reference therein).
Let $m=n/2$. For any permutation $\sigma$ of $1,\ldots,n$, let
$(i_1,\ldots,i_n):=\sigma(1,\ldots,n)$. For $r=1,\ldots,m$, we define
$\bS^{\sigma}_r$ and $\hat{\bDelta}^{\sigma}_r$ to be
\[
\bS^{\sigma}_r:=\bS_{i_{2r},i_{2r-1}} \quad\mbox{and}\quad \hat {
\bDelta }^{\sigma}_r:=\frac{1}{m}\bigl(
\bS^{\sigma}_r\bigl[\bS^{\sigma}_r
\bigr]^T-\Tb\bigr).
\]
Moreover, for $i=1,\ldots,m$, let
\[
\bS_{i}:=\bS_{2i,2i-1}\quad \mbox{and}\quad \hat{\bDelta}_{i}:=
\frac
{1}{m}\bigl(\bS_{i}\bS_{i}^T-\Tb
\bigr).
\]
The convexity of the trace exponential function implies that
%
\begin{eqnarray}
\label{eq:5} \tr \mathrm{e}^{\theta(\hat{\Tb}-\Tb)} & = &\tr \mathrm{e}^{ \theta\sum_{i,i'}\hat{\bDelta}_{i,i'} }
\nonumber
\\
& =& \tr\exp \Biggl\{ \frac{1}{\card(S_n)} \sum_{\sigma\in S_n}
\theta \sum_{r=1}^m \hat{
\bDelta}^{\sigma}_r \Biggr\}
\\
& \le&\frac{1}{\card(S_n)} \sum_{\sigma\in S_n} \tr
\mathrm{e}^{ \theta\sum
_{r=1}^m \hat{\bDelta}_{r}^{\sigma} }\nonumber,
\end{eqnarray}
where $S_n$ is the permutation group of $\{1,\ldots,n\}$. Taking
expectation on both sides of equation~\eqref{eq:5} gives that
%
\begin{equation}
\label{eq:decouple} \E\tr \mathrm{e}^{\theta(\hat{\Tb}-\Tb)} \le\E\tr \mathrm{e}^{ \theta\sum_{i=1}^m \hat{\bDelta}_{i} }.
\end{equation}
According to the definition, $\hat{\bDelta}_1,\ldots, \hat{\bDelta}_m$
are $m$ independent and identically distributed random matrices, and
this finishes the decoupling step.

Combing equations \eqref{eq:LapMethodMatrix} and \eqref{eq:decouple},
we have
%
\begin{equation}
\label{eq: Bernstein1} \P\bigl[ \lambda_{\max} ( \hat{\Tb}-\Tb) \ge t\bigr] \le
\inf_{\theta> 0}\mathrm{e}^{-\theta t} \E\tr \mathrm{e}^{ \theta\sum_{i=1}^m
\hat
{\bDelta}_{i} }.
\end{equation}
Recall that $\E\hat{\bDelta}_{i} = 0$. Following the proof of Theorem~6.1 in Tropp \cite{tropp2011user}, if we can show that there are some
nonnegative numbers $R_1$ and $R_2$ such that
\[
\lambda_{\max}(\hat{\bDelta}_{i} ) \le R_1,\qquad \Biggl\|
\sum_{i=1}^m \E\hat{\bDelta}_i^2
\Biggr\|_2 \le R_2,
\]
then the right-hand side of equation \eqref{eq: Bernstein1} can be
bounded by
\[
\inf_{\theta> 0}\mathrm{e}^{-\theta t} \E\tr \mathrm{e}^{ \theta\sum_{i=1}^m \hat
{\bDelta}_{i} } \leq d \exp
\biggl\{ -\frac{ t^2/2 }{ R_2 + R_1 t/3} \biggr\}.
\]
We first show that $R_1 = \frac{2d}{m}$. Because $\|\hat{\bDelta
}_i\|_{\max}\leq2/m$, by simple calculation, we have
\[
\lambda_{\max}(\hat{\bDelta}_i)\leq\|\hat{
\bDelta}_i\|_1\leq d\cdot \|\hat{\bDelta}_i
\|_{\max}\leq\frac{2d}{m}.
\]
We then calculate $R_2$. For this, we have, because $\bX$ is continuous,
\begin{eqnarray*}
\sum_{i=1}^m \E\hat{
\bDelta}_i^2 = \frac{1}{m}\E\bigl(
\bS_1\bS _1^T-\Tb \bigr)^2=
\frac{1}{m}\bigl(\E\bigl(d\bS_1\bS_1^T
\bigr)-\Tb^2\bigr)=\frac{1}{m}\bigl(d\Tb-\Tb^2
\bigr).
\end{eqnarray*}
Accordingly,
\[
\Biggl\|\sum_{i=1}^m \E\hat{
\bDelta}_i^2\Biggl\|_2\leq\frac{1}{m}\bigl(d
\|\Tb \| _2+\|\Tb\|_2^2\bigr),
\]
so we set $R_2 = \frac{1}{m}(d\|\Tb\|_2+\|\Tb\|_2^2)$.\vspace*{1pt}

Thus, using Theorem~6.1 in Tropp \cite{tropp2011user}, for any
\[
t\leq R_2/R_1=\frac{d\|\Tb\|_2+\|\Tb\|_2^2}{2d},
\]
we have
\[
\P \bigl\{ \lambda_{\max} ( \hat{\Tb}-\Tb) \ge t \bigr\} \leq d\cdot
\exp \biggl( -\frac{3nt^2}{16(d\|\Tb\|_2+\|\Tb\|_2^2)} \biggr).
\]
A similar argument holds for $\lambda_{\max}( - \hat{\Tb} + \Tb)$.
Accordingly, we have
\[
\P \bigl\{\|\hat{\Tb} - \Tb\|_2 \geq t \bigr\} \leq2d\cdot\exp \biggl( -
\frac{3nt^2}{16(d\|\Tb\|_2+\|\Tb\|_2^2)} \biggr).
\]
Finally, when
\[
n\geq\frac{64d^2\log(d/\alpha)}{3(d\|\Tb\|_2+\|\Tb \|_2^2)},
\]
we have
\[
\sqrt{\frac{16(d\|\Tb\|_2+\|\Tb\|_2^2)\log(d/\alpha
)}{3n}}\leq \frac{d\|\Tb\|_2+\|\Tb\|_2^2}{2d}.
\]
This completes the proof.
\end{pf}

\subsection{Proof of Theorem \texorpdfstring{\protect\ref{thm:lowd_spectral}}{3.2}}\label{sec5.2}

To prove Theorem~\ref{thm:lowd_spectral}, we first need the following
lemma, which connects $\sqrt{1-\bSigma_{jk}^2}$ to a Gaussian
distributed random vector $(X,Y)^T\in\reals^2$ and plays a key role in
bounding $\|\hat{\bSigma}-\bSigma\|_2$ by $\|\hat{\Tb }-\Tb\|_2$.
%
\begin{lemma}\label{lemma:E|xy|}
Provided that
\[
\pmatrix{ X
\cr
Y }\sim N_2\left(\zero,\left[
\matrix{ 1 & \sigma
\cr
\sigma& 1 }
 \right] \right),
\]
we have
\[
\E|XY| = \E XY \E\sign(XY)+\frac{2}{\pi}\sqrt{1-\sigma^2}.
\]
\end{lemma}
\begin{pf}
We recall that $\sigma:= \sin( \frac{\pi}{2} \tau)$ with $\tau$ the
Kendall's tau correlation coefficient of $X,Y$.
Without loss of generality, assume that $\sigma> 0, \tau> 0$
(otherwise show for $-Y$ instead of $Y$). Define
\[
\beta_+ = \E|XY| I( XY >0),\qquad \beta_- = \E|XY| I( XY < 0),
\]
where $I(\cdot)$ is the indicator function. We then have
%
\begin{equation}
\label{eqn: betaplusmius} \E|XY| = \beta_+ + \beta_-,\qquad \E XY = \sigma= \beta_+ - \beta_-.
\end{equation}
To compute $\beta_+$, using the fact that
\[
X \stackrel{d} {=} \sqrt{ \frac{1+\sigma}{2} }Z_1 + \sqrt{
\frac
{1-\sigma
}{2} }Z_2,\qquad Y \stackrel{d} {=} \sqrt{ \frac{1+\sigma}{2}
}Z_1 - \sqrt{ \frac
{1-\sigma}{2} }Z_2,
\]
where $Z_1, Z_2 \sim N_1(0,1)$ are independently and identically distributed.

Let $F_{X,Y}$ and $F_{Z_{1}, Z_{2}}$ be the joint distribution
functions of $(X,Y)^{T}$ and $(Z_{1}, Z_{2})^{T}$. We have
\begin{eqnarray*}
\beta_+ &=& \int_{xy>0} |xy| \,\mathrm{d} F_{X,Y}(x,y)
\\
&=& \int_{xy>0} \frac{(x+y)^2-(x-y)^2}{4} \,\mathrm{d} F_{X,Y}(x,y)
\\
&=& \int_{ z_1^2 > ({(1-\sigma)}/{(1+\sigma)}) z_2^2} \biggl( \frac
{1+\sigma}{2}z_1^2
- \frac{1-\sigma}{2}z_2^2 \biggr) \,\mathrm{d} F_{Z_1,Z_2}(z_1,z_2)
\\
&=& \int_0^{+\infty} \int_{-\alpha}^\alpha
2 \biggl\{ \frac{1+\sigma}{2}r^2\cos^2(\theta) -
\frac{1-\sigma
}{2}r^2\sin^2(\theta) \biggr\}\cdot
\frac{1}{2\pi} \mathrm{e}^{-r^2/2} r \,\mathrm{d}\theta \,\mathrm{d}r,
\end{eqnarray*}
where $\alpha:= \arcsin (\sqrt{\frac{1+\sigma}{2}} ) $. By
simple calculation, we have
\[
\int_{0}^{\infty}r^3\mathrm{e}^{-r^2/2}\,\mathrm{d}r=
\frac{1}{2}\int_0^{\infty}ue^{-u/2}\,\mathrm{d}u=2.
\]
Accordingly, we can proceed the proof and show that
%
\begin{eqnarray}
\label{eqn:beta+} \beta_+ &=& \int_0^{+\infty} \int
_{-\alpha}^\alpha \bigl(\cos(2\theta)+\sigma\bigr)\cdot
r^3\frac{1}{2\pi} \mathrm{e}^{-r^2/2} \,\mathrm{d}\theta \,\mathrm{d}r
\nonumber
\\[-8pt]
\\[-8pt]
\nonumber
&=&\frac{1}{\pi} \bigl(\sin(2\alpha)+2\alpha\sigma \bigr).
\end{eqnarray}
Since $\sin(2\alpha) = \sqrt{1-\sigma^2}=\cos(\pi\tau/2)$ and
$\alpha
\geq\arcsin(\sqrt{1/2})\geq\pi/4$, we have that $2 \alpha= \frac
{\pi
}{2}(1+\tau)$, and then equation \eqref{eqn:beta+} continues to give
\[
\beta_+ = \frac{\sigma}{2}(1+\tau) + \frac{1}{\pi}\sqrt{1-
\sigma^2}.
\]
Combined with equation \eqref{eqn: betaplusmius} gives the equality claimed.
\end{pf}

Using Theorem~\ref{thm:lowd_spectral2} and Lemma~\ref{lemma:E|xy|}, we
proceed to prove Theorem~\ref{thm:lowd_spectral}.

\begin{pf*}{Proof of Theorem~\ref{thm:lowd_spectral}}
Using Taylor
expansion, for any $j\ne k$, we have
\[
\sin \biggl(\frac{\pi}{2}\hat{\tau}_{jk} \biggr)-\sin \biggl(
\frac
{\pi
}{2}\tau_{jk} \biggr) =\cos \biggl(\frac{\pi}{2}
\tau_{jk} \biggr)\frac{\pi}{2}(\hat {\tau }_{jk}-
\tau_{jk}) -\frac{1}{2}\sin(\theta_{jk}) \biggl(
\frac{\pi}{2} \biggr)^{2}(\hat {\tau }_{jk}-
\tau_{jk})^{2},
\]
where $\theta_{jk}$ lies between $\tau_{jk}$ and $\hat{\tau}_{jk}$.
Thus,
\[
\hat{\bSigma}-\bSigma=\Eb_{1}+\Eb_{2},
\]
where $\Eb_1,\Eb_2\in\reals^{d\times d}$ satisfy that for $j\neq k$,
\begin{eqnarray*}
[\Eb_{1}]_{jk} & =& \cos \biggl(\frac{\pi}{2}
\tau_{jk} \biggr)\frac
{\pi
}{2}(\hat{\tau}_{jk}-
\tau_{jk}),
\\
{} [\Eb_{2}]_{jk} & =& -\frac{1}{2}\sin(
\theta_{jk}) \biggl(\frac{\pi
}{2} \biggr)^{2}(\hat{
\tau}_{jk}-\tau_{jk})^{2},
\end{eqnarray*}
and the diagonal entries of both $E_{1}$ and $E_{2}$ are all zero.

Using the results of $U$-statistics shown in Hoeffding \cite
{hoeffding1963probability}, we have that for any $j\ne k$ and $t>0$,
\[
\P\bigl(|\hat{\tau}_{jk}-\tau_{jk}|>t\bigr)<2\mathrm{e}^{-nt^{2}/4}.
\]
For some constant $\alpha$, let the event $\Omega_{2}$ be defined as
\[
\Omega_{2}:= \biggl\{\exists1\le j\ne k\le d,\bigl |[\Eb_{2}]_{jk}\bigr|>
\pi ^{2}\cdot\frac{\log(d/\alpha)}{n} \biggr\}.
\]
Since $ |[\Eb_{2}]_{jk} |\le\frac{\pi^{2}}{8}(\hat{\tau
}_{jk}-\tau_{jk})^{2}$,
by union bound, we have
\[
\P(\Omega_{2}) \leq\frac{d^{2}}{2}\cdot2\mathrm{e}^{-2\log(d/\alpha)} =
\alpha^2.
\]
Conditioning on $\Omega_{2}^C$, for any $\bv\in\S^{d-1}$, we have
%
\begin{eqnarray}
\label{eq:E2} \bigl\llvert \bv^{T} \Eb_{2} \bv\bigr\rrvert
\le\sqrt{\sum_{j,k\in J}[\Eb
_{2}]_{jk}^{2}}\cdot\|\bv\|_2^2
\leq\sqrt{d^2 \biggl(\pi ^{2}\cdot \frac{\log(d/\alpha)}{n}
\biggr)^2}= \pi^2\cdot\frac{d\log
(d/\alpha)}{n}.
\end{eqnarray}

We then analyze the term $\Eb_1$. Let $\Wb=[\Wb_{jk}]\in\reals
^{d\times
d}$ with
$\Wb_{jk}=\frac{\pi}{2}\cos(\frac{\pi}{2}\tau_{jk})$ and $\hat
{\Tb
}=[\hat{\Tb}_{jk}]$ be the Kendall's
tau sample correlation matrix with $\hat{\Tb}_{jk}=\hat{\tau
}_{jk}$. We
can write
\[
\Eb_1=\Wb\circ(\hat{\Tb}-\Tb),
\]
where $\circ$ represents the Hadamard product. Given the spectral norm
bound of $\hat{\Tb}-\Tb$ shown in Theorem~\ref{thm:lowd_spectral2}, we
now focus on controlling $\Eb_1$. 
Let $\bY:=(Y_1,\ldots,Y_d)^T\sim N_d(\zero,\bSigma)$ follow a Gaussian
distribution with mean zero and covariance matrix $\bSigma$. Using the
equality in Lemma~\ref{lemma:E|xy|}, we have, for any $j\ne k$,
\[
\E|Y_jY_k| = \tau_{jk}\bSigma_{jk}
+ \frac{2}{\pi}\sqrt{1 - \bSigma_{jk}^2}.
\]
Reminding that
\[
\cos \biggl(\frac{\pi}{2}\tau_{jk} \biggr)=\sqrt{1-
\sin^2 \biggl(\frac{\pi
}{2}\tau_{jk} \biggr)}=\sqrt
{1-\bSigma_{jk}^2},
\]
we have
\[
\Wb_{jk}=\frac{\pi}{2}\cos \biggl(\frac{\pi}{2}
\tau_{jk} \biggr)=\frac{\pi
^2}{4}\bigl(\E|Y_jY_k|-
\tau_{jk}\bSigma_{jk}\bigr).
\]
Then let $\bY':=(Y_1',\ldots,Y_d')^T\in\reals^d$ be an independent copy
of $\bY$. We have, for any $\bv\in\S^{d-1}$ and symmetric matrix
$\Mb\in
\reals^{d\times d}$,
\begin{eqnarray}
\label{eq:6} \bigl| \bv^T \Mb\circ\Wb\bv\bigr| &=&\Biggl | \sum
_{j,k=1}^d v_j v_k
\Mb_{jk} \Wb_{jk} \biggr|
\nonumber
\\
&=& \biggl\llvert \E\frac{\pi^2}{4} \sum_{j,k}
v_j v_k \Mb_{jk} \bigl(|Y_jY_k|
- Y_j Y_k \sign\bigl(Y_j'
Y_k'\bigr)\bigr) \biggr\rrvert
\nonumber
\\
&\le&\frac{\pi^2}{4} \E \biggl( \biggl\llvert \sum
_{j,k} v_j v_k \Mb_{jk}
|Y_jY_k| \biggr\rrvert + \biggl\llvert \sum
_{j,k} v_j v_k \Mb_{jk}
Y_j Y_k \sign\bigl(Y_j'
Y_k'\bigr) \biggr\rrvert \biggr)
\nonumber
\\[-8pt]
\\[-8pt]
\nonumber
&\le&\frac{\pi^2}{4} \|\Mb\|_2 \cdot\E \biggl(2 \sum
_j v_j^2 Y_j^2
\biggr)
\nonumber
\\
&=& \frac{\pi^2}{4} \|\Mb\|_2 \cdot \biggl(2 \sum
_j v_j^2 \biggr)
\nonumber
\\
&=& \frac{\pi^2}{2} \|\Mb\|_2.\nonumber
\end{eqnarray}
Here, the second inequality is due to the fact that for any $\Mb\in
\reals^{d\times d}$ and $\bv\in\reals^d$, $|\bv^T\Mb\bv|\leq
\|\Mb \|_2\|\bv\|_2$ and the third equality is due to the fact that
$\E
Y_j^2=\bSigma_{jj}=1$ for any $j\in\{1,\ldots,d\}$. Accordingly, we have
%
\begin{equation}
\label{eq:E1} \|\Eb_1\|_2=\bigl\|\Wb\circ(\hat{\Tb}-\Tb)
\bigr\|_2\leq\frac{\pi^2}{2} \|\hat{\Tb}-\Tb\|_2.
\end{equation}
The bound in Theorem~\ref{thm:lowd_spectral}, with $\bSigma$ being
replaced by $\Tb$, follows from the fact that
\[
\|\hat{\bSigma}-\bSigma\|_2=\|\Eb_1+\Eb_2
\|_2\leq\|\Eb _1\|_2+\| \Eb_2
\|_2
\]
and by combining equations \eqref{eq:TB}, \eqref{eq:E2} and \eqref
{eq:E1}. Finally, we prove that $\|\Tb\|_2\leq\|\bSigma \|_2$. We
have $\Tb_{jk}=\frac{2}{\pi}\arcsin(\bSigma_{jk})$. Using the Taylor
expansion and the fact that $|\bSigma_{jk}|\leq1$ for any $(j,k)\in\{
1,\ldots,d\}$, we have
\[
\Tb=\frac{2}{\pi}\sum_{m=0}^{\infty}
\frac
{(2m)!}{4^m(m!)^2(2m+1)}\underbrace{(\bSigma\circ\cdots\circ \bSigma)}_{2m+1}.
\]
By Schur's theorem (see, e.g., page 95 in Johnson \cite
{johnson1990matrix}), we
have for any two positive semidefinite matrices $\Ab$ and $\Bb$,
\[
\|\Ab\circ\Bb\|_2\leq\Bigl(\max_j
\Ab_{jj}\Bigr)\|\Bb\|_2.
\]
Accordingly, using the fact that $\bSigma_{jj}=1$ for all $1\leq j\leq
d$, we have
\[
\bigl\|\underbrace{(\bSigma\circ\cdots\circ\bSigma )}_{2m+1}\bigr\|_2
\leq\| \bSigma\|_2,
\]
implying that
%
\begin{eqnarray}
\label{eq:T2lessS2} \|\Tb\|_2&\leq&\|\bSigma\|_2\cdot
\frac{2}{\pi}\sum_{m=0}^{\infty}
\frac{(2m)!}{4^m(m!)^2(2m+1)}
\nonumber
\\[-8pt]
\\[-8pt]
\nonumber
&=&\|\bSigma\|_2\cdot\frac{2}{\pi
}\arcsin 1=\|
\bSigma\|_2.
\end{eqnarray}
Accordingly, we can replace $\Tb$ with $\bSigma$ in the upper bound and
have the desired result.
\end{pf*}

\subsection{Proofs of Propositions \texorpdfstring{\protect\ref{prop:example1}}{4.5} and
\texorpdfstring{\protect\ref{prop:example2}}{4.6}}\label{sec5.3}

Proposition~\ref{prop:example1} is a direct consequence of Lemma~\ref
{lem:subgaussian}. To prove Proposition~\ref{prop:example2}, we first
introduce the subexponential norm.
For any random variable $X\in\reals$, $\|X\|_{\phi_1}$ is defined
as follows:
\[
\|X\|_{\phi_1}:=\sup_{k\geq1} \frac{1}{k} \bigl(
\E|X|^k \bigr)^{1/k}.
\]
Let $\bS:=\sign(\bX-\tilde{\bX})$. Because $\bv^T\bS$ is sub-Gaussian
and $\E\bv^T\bS=0$, using Lemma~5.14 in Vershynin \cite
{vershynin2010introduction}, we get
\begin{eqnarray*}
\bigl\|\bigl(\bv^T\bS\bigr)^2-\E\bigl(\bv^T\bS
\bigr)^2\bigr\|_{\phi_1}&\leq&\bigl\|\bigl(\bv ^T\bS
\bigr)^2\bigr\| _{\phi_1}+\bigl\|\bv^T\Tb\bv
\bigr\|_{\phi_1}
\\
&\leq&2\bigl\|\bv^T\bS\bigr\|_{\phi_2}^2+\bv^T
\Tb\bv
\\
&\leq&(L_2+1)\|\Tb\|_2. 
\end{eqnarray*}
Since $(\bv^T\bS)^2-\E(\bv^T\bS)^2$ is a zero-mean random variable and
$\bv^T\bS$ is sub-Gaussian, using Lemma~5.15 in Vershynin \cite
{vershynin2010introduction}, there exist two fixed constants $C',c'$
such that if $|t|\leq c'/\|(\bv^T\bS)^2-\E(\bv^T\bS)^2\|_{\phi_1}$,
we have
\[
\E\exp\bigl(t\bigl(\bigl(\bv^T\bS\bigr)^2-\E\bigl(
\bv^T\bS\bigr)^2\bigr)\bigr)\leq\exp \bigl(C't^2
\bigl\|\bigl(\bv ^T\bS\bigr)^2-\E\bigl(\bv^T\bS
\bigr)^2\bigr\|_{\phi_1}^2 \bigr).
\]
Accordingly, by choosing $t_0=c'(L_2+1)^{-1}\|\Tb\|_2^{-1}$ and
$K=C'(L_2+1)^2$ in equation \eqref{eq:subgaussian}, noticing that
$t_0\|\Tb\|_2=c'(L_2+1)^{-1}$,
the sign sub-Gaussian condition is satisfied.

\subsection{Proof of Theorem \texorpdfstring{\protect\ref{thm:example}}{4.7}}\label{sec:example1}

In this section, we provide the proof of Theorem~\ref{thm:example}. In
detail, we show that for any transelliptically distributed random
vector $\bX$ such that $f(\bX)\sim \mathit{EC}_d(\zero,\Ib_d,\xi)$, we have that
$\bX$ satisfies the condition in equation \eqref{eq:subgaussian}.

\begin{pf}
Because for any strictly increasing function $g\dvtx\reals\rightarrow
\reals
$ and $x,y\in\reals$, $\sign(g(x)-g(y))=\sign(x-y)$, $\sign(\xi
x)=\sign
(x)$ (a.s.) for any $\xi$ with $\P(\xi>0)=1$, and the fact that the
elliptical family is closed to the independent sums (Lindskog et al.
\cite
{lindskog2003kendall}), we only need to consider the random vector $\bX
\sim N_d(\zero,\Ib_d)$. For $\bX=(X_1,\ldots,X_d)^T\sim N_d(\zero
,\Ib
_d)$ and $\tilde{\bX}$ as an independent copy of $\bX$, we have $\bX
-\tilde{\bX}\sim N_d(\zero,2\Ib_d)$. Reminding that the off-diagonal
entries of $\Ib_d$ are all zero, defining $\bX^0=(X_1^0,\ldots
,X_d^0)^T=\bX-\tilde\bX$ and
\[
g\bigl(\bX^0,\bv\bigr):=\sum_{j,k}v_jv_k
\sign\bigl(X_j^0X_k^0\bigr),
\]
we have
\[
\bigl\{\bv^T\sign(\bX-\tilde{\bX})\bigr\}^2-\E\bigl\{
\bv^T\sign(\bX-\tilde {\bX})\bigr\} ^2=g\bigl(
\bX^0,\bv\bigr)-\E g\bigl(\bX^0,\bv\bigr).
\]
Accordingly, to bound $\psi (\langle\sign(\bX-\tilde{\bX
}),\bv \rangle;2 )$, we only need to focus on $g(\bX^0,\bv)$.
Letting $\bS
:=(S_1,\ldots,S_d)^T$ with $S_j:=\sign(Y_j^0)$ for $j=1,\ldots,d$.
Using the property of Gaussian distribution, $S_1,\ldots,S_d$ are
independent Bernoulli random variables in $\{-1,1\}$ almost surely. We
then have
\[
g\bigl(\bY^0,\bv\bigr)-\E g\bigl(\bY^0,\bv\bigr)=\sum
_{j,k}v_jv_k\sign
\bigl(Y_j^0Y_k^0\bigr)-1=\bigl(
\bv ^T\bS\bigr)^2 - 1.
\]
Here, the first equality is due to the fact that $\|\bv\|_2=\sum_{j=1}^d v_j^2 = 1$.

We then proceed to analyze the property of $(\bv^T\bS)^2 - 1$. By the
Hubbard--Stratonovich transform (Hubbard \cite
{hubbard1959calculation}), for any
$\eta\in\reals$,
%
\begin{equation}
\label{eq:hs_transform} \exp\bigl(\eta^2\bigr) = \int_{-\infty}^{\infty}
\frac{1}{\sqrt{4\pi}} \mathrm{e}^{-y^2/4
+ y\eta} \,\mathrm{d}y.
\end{equation}
Using equation \eqref{eq:hs_transform}, we have that, for any $t>0$,
\begin{eqnarray*}
\E\exp\bigl[t\bigl\{\bigl(\bv^T\bS\bigr)^2 - 1\bigr\}
\bigr]&= & \mathrm{e}^{-t}\E \mathrm{e}^{t(\bv^{T}\bS)^{2}}
\\
&= & \frac{\mathrm{e}^{-t}}{\sqrt{4\pi t}}\int_{-\infty}^{+\infty
}\mathrm{e}^{-y^{2}/4t}
\mathbb{E}\mathrm{e}^{ y\sum_{j=1}^d v_{j}S_j }\,\mathrm{d}y
\\
&= & \frac{\mathrm{e}^{-t}}{\sqrt{4\pi t}}\int_{-\infty}^{+\infty
}\mathrm{e}^{-y^{2}/4t}
\prod_{j=1}^d \frac{1}{2}
\bigl(\mathrm{e}^{yv_{j}}+\mathrm{e}^{-yv_{j}}\bigr)\,\mathrm{d}y.
\end{eqnarray*}
For any number $z\in\mathbb{N}$, we define $z!$ to represent the
factorial product of $z$. Because for any $a\in\reals$, by Taylor
expansion, we have
\[
\bigl\{\exp(a)+\exp(-a)\bigr\}/2 = \sum_{k=0}^{\infty}
a^{2k}/(2k)! \quad\mbox {and}\quad \exp\bigl(a^2/2\bigr)=\sum
_{k=0}^{\infty}a^{2k}/
\bigl(2^k\cdot k!\bigr).
\]
Because $(2k)!>2^k\cdot k!$, we have
\[
\bigl\{\exp(a)+\exp(-a)\bigr\}/2 \leq\exp\bigl(a^2/2\bigr).
\]
Accordingly, we have for any $0<t<1/4$,
\begin{eqnarray*}
\E\exp\bigl[t\bigl\{\bigl(\bv^T\bS\bigr)^2 - 1\bigr\}
\bigr]& = & \frac{\mathrm{e}^{-t}}{\sqrt{4\pi
t}}\int_{-\infty}^{+\infty}\mathrm{e}^{-y^{2}/4t}
\prod_{j=1}^d \frac
{1}{2}
\bigl(\mathrm{e}^{yv_{j}}+\mathrm{e}^{-yv_{j}}\bigr)\,\mathrm{d}y
\\
&\le& \frac{\mathrm{e}^{-t}}{\sqrt{4\pi t}}\int_{-\infty}^{+\infty
}\mathrm{e}^{-y^{2}/4t}\mathrm{e}^{\sum_{j=1}^d({1}/{2})y^{2}v_{j}^{2}}\,\mathrm{d}y
\\
&= & \frac{\mathrm{e}^{-t}}{\sqrt{4\pi t}}\int_{-\infty}^{+\infty
}\mathrm{e}^{-y^{2}/4t+({1}/{2})y^{2}}\,\mathrm{d}y
\\
&= & \frac{\mathrm{e}^{-t}}{\sqrt{1-2t}}.
\end{eqnarray*}
By Taylor expansion of $\log(1-x)$, we have that
\[
\frac{1}{\sqrt{1-2t}}=\exp \Biggl\{\frac{1}{2}\sum
_{k=1}^{\infty
}\frac
{(2t)^k}{k} \Biggr\},
\]
which implies that for all $0<t<1/4$,
\[
\frac{\mathrm{e}^{-t}}{\sqrt{1-2t}}=\exp \Biggl(t^2+\frac{1}{2}\sum
_{k=3}^{\infty
}\frac{(2t)^k}{k} \Biggr)\leq\exp
\bigl(2t^2\bigr).
\]
This concludes that for $0<t<1/4$,
%
\begin{equation}
\label{eq:1} \E\exp\bigl[t\bigl\{\bigl(\bv^T\bS\bigr)^2
- 1\bigr\}\bigr] \leq\exp\bigl(2t^2\bigr).
\end{equation}
Due to that $(\bv^{T}\bS)^{2} \ge0$, we can apply Theorem~2.6 in
Chung and Lu \cite
{chung2006complex} to control the term $\E\exp[t\{1-(\bv^T\bS)^2\}]$.
In detail, suppose that the random variable $Y$ satisfying $\mathbb
{E}Y=0$, $Y\le a_{0}$, and $\mathbb{E}Y^{2}=b_{0}$ for some absolute
constants $a_0$ and $b_0$.
Then for any $0<t<2/a_0$, using the proof of Theorem~2.8 in Chung and
Lu \cite
{chung2006complex}, we have
%
\begin{equation}
\label{eq:4} \mathbb{E}\mathrm{e}^{tY}\le\exp\bigl\{3b_{0}/2\cdot
t^{2}\bigr\}.
\end{equation}
%
For $Y=1-(\bv^{T}\bS)^{2}$, we have
%
\begin{eqnarray}
\label{eq:3} a_0=1\quad \mbox{and}\quad b_{0}=\mathbb{E}\bigl(
\bv^{T}\bS\bigr)^{4}-1=2-2\sum_{j=1}^dv_{j}^{4}<2.
\end{eqnarray}
Here, we remind that $\E(\bv^T\bS)^2=\sum_jv_j^2=1$. Combining
equations \eqref{eq:4} and \eqref{eq:3} implies that for any $t>0$,
%
\begin{eqnarray}
\label{eq:2} \E\exp\bigl[t\bigl\{1-\bigl(\bv^T\bS
\bigr)^2\bigr\}\bigr]\leq\exp\bigl\{3t^2\bigr\}.
\end{eqnarray}
Combining equations \eqref{eq:1} and \eqref{eq:2}, we see that equation
\eqref{eq:subgaussian} holds with $K=3/4$ and $t_0=1/4$ (reminding that
here $\|\Tb\|_2=1$).
\end{pf}

\subsection{Proof of Theorem \texorpdfstring{\protect\ref{thm:example2}}{4.8} and
Corollary \texorpdfstring{\protect\ref{cor:compound}}{4.9}}\label{sec5.5}
In this section, we prove Theorem~\ref{thm:example2} and Corollary~\ref
{cor:compound}. Using the same argument as in the proof of Theorem~\ref
{thm:example},
we only need to focus on those random vectors that are Gaussian distributed.

\begin{pf*}{Proof of Theorem~\ref{thm:example2}}
Assume that $\bSigma\in\reals^{d\times d}$ is a compound symmetric
matrix such that
\[
\bSigma_{jj}=1 \quad\mbox{and}\quad \bSigma_{jk}=\rho\qquad\mbox{for } j
\neq k.
\]
By the discussion on page 11 of Vershynin \cite
{vershynin2010introduction}, to
prove equation \eqref{eq:subgaussian3} holds, we only need to prove
that for $0 \leq\rho\leq C_0$ where $C_0$ is some absolute constant,
$\bX=(X_1,\ldots,X_d)^T\sim N_d(\zero,\bSigma)$ and $\bv\in\S^{d-1}$,
we have
\[
\exp\bigl(t\bv^T\sign(\bX-\tilde{\bX})\bigr) \leq\exp\bigl(c\|\Tb
\|_2t^2\bigr),
\]
for some fixed constant $c$. This result can be proved as follows. Let
$\eta_0,\eta_1,\ldots,\eta_d$ be i.i.d. standard Gaussian random
variables, then $\bZ:=\bX-\tilde{\bX}$ can be expressed as $\bZ
\stackrel
{d}{=}(Z_1',\ldots,Z_d')^T$, where
\begin{eqnarray*}
Z_1'&=&\sqrt{2\rho}\eta_0+\sqrt{2-2\rho}
\eta_1,
\\
Z_2'&=&\sqrt{2\rho}\eta_0+\sqrt{2-2\rho}
\eta_2,
\\
&\cdots&
\\
Z_d'&=&\sqrt{2\rho}\eta_0+\sqrt{2-2\rho}
\eta_d.
\end{eqnarray*}
Accordingly, we have
\begin{eqnarray*}
\E\exp\bigl(t \bv^T\sign(\bX-\tilde{\bX})\bigr)&=&\E \Biggl(\exp
\Biggl(t\sum_{j=1}^dv_j\sign(
\sqrt{2\rho}\eta_0+\sqrt{2-2\rho}\eta_j)\Biggr) \Biggr)
\\
&=&\E \Biggl(\E\Biggl(\exp\Biggl(t\sum_{j=1}^dv_j
\sign(\sqrt{2\rho}\eta _0+\sqrt{2-2\rho }\eta_j)\Biggr) \Big|
\eta_0\Biggr) \Biggr)
\end{eqnarray*}
Moreover, we have
%
\begin{eqnarray}
\label{eq:new3} \sqrt{2\rho}\eta_0+\sqrt{2-2\rho}\eta_j |
\eta_0 \sim N_1(\sqrt {2\rho }\eta_0, 2-2
\rho).
\end{eqnarray}
Letting $\mu:=\sqrt{2\rho}\eta_0$ and $\sigma:=\sqrt{2-2\rho}$,
equation \eqref{eq:new3} implies that
\[
\P(\sqrt{2\rho}\eta_0+\sqrt{2-2\rho}\eta_j>0 |
\eta_0) = \Phi \biggl( \frac
{\mu}{\sigma} \biggr),
\]
where $\Phi(\cdot)$ is the CDF of the standard Gaussian. This further
implies that
\[
\sign(\sqrt{2\rho}\eta_0+\sqrt{2-2\rho}\eta_j)|
\eta_0 \sim \operatorname{Bern}\biggl(\Phi \biggl( \frac{\mu}{\sigma}
\biggr) \biggr),
\]
where we denote $Y\sim\operatorname{Bern}(p)$ if $\P(Y=1)=p$ and $\P
(Y=-1)=1-p$. Accordingly, letting $\alpha:=\Phi(\mu/\sigma)$, we have
\[
\E \bigl(\exp\bigl(tv_j\sign(\sqrt{2\rho}\eta_0+
\sqrt{2-2\rho}\eta _j) \bigr)|\eta _0 \bigr)=(1-
\alpha)\mathrm{e}^{-v_jt}+\alpha \mathrm{e}^{v_jt}.
\]
Letting $\beta:=\alpha-1/2$, we have
\[
\E \bigl(\exp\bigl(tv_j\sign(\sqrt{2\rho}\eta_0+
\sqrt{2-2\rho}\eta _j) \bigr)|\eta _0 \bigr)=
\tfrac{1}{2}\mathrm{e}^{-v_jt}+\tfrac{1}{2} \mathrm{e}^{v_jt}+\beta
\bigl(\mathrm{e}^{v_jt}-\mathrm{e}^{-v_jt}\bigr).
\]
Using that fact that $\frac{1}{2}\mathrm{e}^{a}+\frac{1}{2} \mathrm{e}^{-a}\leq
\mathrm{e}^{a^2/2}$, we have
\[
\E \bigl(\exp\bigl(tv_j\sign(\sqrt{2\rho}\eta_0+
\sqrt{2-2\rho}\eta _j) \bigr)|\eta _0 \bigr)\leq\exp
\bigl(v_j^2t^2/2\bigr)+\beta
\bigl(\mathrm{e}^{v_jt}-\mathrm{e}^{-v_jt}\bigr).
\]
Because conditioning on $\eta_0$, $\sign(\sqrt{2\rho}\eta_0+\sqrt
{2-2\rho}\eta_j)$, $j=1,\ldots,d$, are independent of each other, we have
\begin{eqnarray*}
&&\E \Biggl(\exp \Biggl(t\sum_{j=1}^dv_j
\sign(\sqrt{2\rho}\eta _0+\sqrt {2-2\rho}\eta_j) \Biggr)\bigg|
\eta_0 \Biggr)\\
&&\quad\leq \prod_{j=1}^d
\bigl\{\exp\bigl(v_j^2t^2/2\bigr)+\beta
\bigl(\mathrm{e}^{v_jt}-\mathrm{e}^{-v_jt}\bigr) \bigr\}
\\
&&\quad=\mathrm{e}^{t^2/2} \Biggl(1+\sum_{k=1}^d
\beta^k\sum_{j_1<j_2<\cdots<j_k} \prod
_{j\in\{j_1,\ldots,j_k\}}\frac
{\mathrm{e}^{v_jt}-\mathrm{e}^{-v_jt}}{\mathrm{e}^{v_j^2t^2/2}} \Biggr).
\end{eqnarray*}
Moreover, for any centered Gaussian distribution $Y\sim N_1(0,\kappa)$
and $t\in\reals$, we have
\begin{eqnarray*}
\P\bigl(\Phi(Y)>1/2+t\bigr)&=&\P\bigl(Y>\Phi^{-1}(1/2+t)\bigr)
\\
&=&\P\bigl(Y>-\Phi^{-1}(1/2-t)\bigr)
\\
&=&\P\bigl(Y<\Phi^{-1}(1/2-t)\bigr)
\\
&=&\P\bigl(\Phi(Y)<1/2-t\bigr).
\end{eqnarray*}
Combined with the fact that $\Phi(Y)\in[0,1]$, we have
\[
\E\bigl(\Phi(Y)-1/2\bigr)^k=0 \qquad\mbox{when } k \mbox{ is odd}.
\]
This implies that when $k$ is odd,
\[
\E\beta^k=0 = \E \bigl(\Phi\bigl(\sqrt{\rho/(1-\rho)}
\eta_0\bigr)-\tfrac
{1}{2} \bigr)^k=0.
\]
Accordingly, denoting $\varepsilon=\E\exp (t\sum_{j=1}^dv_j\sign
(\sqrt
{2\rho}\eta_0+\sqrt{2-2\rho}\eta_j)  )$, we have
\[
\varepsilon\leq \mathrm{e}^{t^2/2} \biggl(1+\sum_{k\ \mathrm{is\ even}}
\E \beta ^k\sum_{j_1<j_2<\cdots<j_k} \prod
_{j\in\{j_1,\ldots,j_k\}}\frac
{\mathrm{e}^{v_jt}-\mathrm{e}^{-v_jt}}{\mathrm{e}^{v_j^2t^2/2}} \biggr).
\]
Using the fact that
\begin{eqnarray*}
\bigl\llvert \mathrm{e}^{a}-\mathrm{e}^{-a}\bigr\rrvert &=&\Biggl\llvert \sum
_{j=1}^{\infty} \frac{a^j}{j!} -\sum
_{j=1}^{\infty} \frac{(-a)^j}{j!}\Biggr\rrvert
\\
&=&2\Biggl\llvert \sum_{m=0}^{\infty}
\frac{a^{2m+1}}{(2m+1)!}\Biggr\rrvert
\\
&=&2|a|\cdot\Biggl\llvert \sum_{m=0}^{\infty}
\frac{a^{2m}}{(2m+1)!}\Biggr\rrvert
\\
&\leq&2|a|\exp\bigl(a^2/2\bigr),
\end{eqnarray*}
we further have
\begin{eqnarray*}
\varepsilon&\leq& \mathrm{e}^{t^2/2} \biggl(1+\sum_{k\ \mathrm{is\ even}}
\E \beta ^k\sum_{j_1<j_2<\cdots<j_k} \prod
_{j\in\{j_1,\ldots,j_k\}}2|v_jt| \biggr)
\\
&=& \mathrm{e}^{t^2/2} \biggl(1+\sum_{k\ \mathrm{is\ even}}\E\beta
^k\bigl(2|t|\bigr)^k\sum_{j_1<j_2<\cdots<j_k}|v_{j_1}
\cdots v_{j_k}| \biggr).
\end{eqnarray*}
By Maclaurin's inequality, for any $x_1,\ldots,x_d\geq0$, we have
\[
\frac{x_1+\cdots+x_n}{n} \geq \biggl(\frac{\sum_{1\leq i<j\leq
n}x_ix_j}{{n\choose 2}} \biggr)^{1/2} \geq
\cdots\geq(x_1\cdots x_n)^{1/n}.
\]
Accordingly,
%
\begin{eqnarray}
\label{eq:new1} &&\mathrm{e}^{t^2/2} \biggl(1+\sum_{k\ \mathrm{is\ even}}
\E\beta^k\bigl(2|t|\bigr)^k\sum_{j_1<j_2<\cdots<j_k}|v_{j_1}
\cdots v_{j_k}| \biggr)
\nonumber
\\
&&\quad\leq \mathrm{e}^{t^2/2} \biggl(1+\sum_{k\ \mathrm{is\ even}} \E\beta
^k\bigl(2|t|\bigr)^k \biggl\{ \pmatrix{n\cr 2}\cdot\bigl(\|\bv\|_1/d\bigr)^k\biggr\} \biggr)
\\
&&\quad\leq \mathrm{e}^{t^2/2} \biggl(1+\sum_{k\ \mathrm{is\ even}} \E
\beta^k\bigl(2|t|\bigr)^k d^{k/2}(e/k)^k
\biggr).\nonumber
\end{eqnarray}
The last inequality is due to the fact that $\|\bv\|_1\leq\sqrt
{d}\|\bv\|_2=\sqrt{d}$ and ${n\choose 2}\leq(ed/k)^k$.

Finally, we analyze $\E\beta^{2m}$ for $m=1,2,\ldots.$ Reminding that
\[
\beta:= \Phi \biggl(\sqrt{\frac{\rho}{1-\rho}}\eta_0 \biggr) -
\frac{1}{2},
\]
consider the function $f(x)\dvtx x\rightarrow\Phi(\sqrt{\rho
/(1-\rho)}x)$,
we have
\[
\bigl|f'(x)\bigr|=\sqrt{\frac{\rho}{1-\rho}}\cdot\frac{1}{\sqrt{2\pi
}}\exp
\biggl(-\frac{\rho}{2(1-\rho)}x^2 \biggr)\leq\sqrt{\frac{\rho}{2\pi
(1-\rho)}}.
\]
Accordingly, $f(\cdot)$ is a Lipschitz function with a Lipschitz
constant $K_0:=\sqrt{\frac{\rho}{2\pi(1-\rho)}}$. By the concentration
of Lipschitz functions of Gaussian (Ledoux \cite
{ledoux2001concentration}), we have
\[
\P\bigl(|\beta|>t\bigr)=\P\bigl(\bigl|f(\eta_0)-\E f(\eta_0)\bigr|>t\bigr)
\leq2\exp\bigl(-t^2/\bigl(2K_0^2\bigr)\bigr).
\]
This implies that, for $m=1,2,\ldots,$
\begin{eqnarray*}
\E\beta^{2m} &=& 2m \int_{0}^{\infty}
t^{2m-1}\P\bigl(|\beta|>t\bigr)\,\mathrm{d}t
\\
&\leq&4m\int_0^{\infty} t^{2m-1}\exp
\bigl(-t^2/\bigl(2K_0^2\bigr)\bigr)\,\mathrm{d}t
\\
&=&4m(\sqrt{2}K_0)^{2m} \int_0^{\infty}t^{2m-1}
\exp\bigl(-t^2\bigr)\,\mathrm{d}t
\\
&=&2m\bigl(2K_0^2\bigr)^m\int
_0^{\infty}t^{m-1}\exp(-t)\,\mathrm{d}t.
\end{eqnarray*}
Using the fact that $\int_0^{\infty}\exp(-t)\,\mathrm{d}t=1$ and for any $m\geq1$,
\[
m\int_0^{\infty}t^{m-1}\exp(-t)\,\mathrm{d}t=\int
_0^{\infty}\exp (-t)\,\mathrm{d}t^m=\int
_0^{\infty}t^m\exp(-t)\,\mathrm{d}t,
\]
we have for $m\in\mathbb{Z}^+$, $\int_0^{\infty}t^{m}\exp(-t)\,\mathrm{d}t=m!$.
Accordingly,
\[
\E\beta^{2m} \leq2m\bigl(2K_0^2
\bigr)^m(m-1)!=2\bigl(2K_0^2
\bigr)^mm!.
\]
Plugging the above result into equation \eqref{eq:new1}, we have
\begin{eqnarray*}
\varepsilon&\leq& \mathrm{e}^{t^2/2}\Biggl(1+\sum_{m=1}^{\infty}
2\bigl(2K_0^2\bigr)^mm!(2t)^{2m}d^m
\bigl(e/(2m)\bigr)^{2m}\Biggr)
\\
&=&\mathrm{e}^{t^2/2}\Biggl(1+\sum_{m=1}^{\infty}
\bigl(K_0^2d\bigr)^m\cdot m!2(2\sqrt
{2}et)^{2m}/(2m)^{2m}\Biggr).
\end{eqnarray*}
Reminding that $\rho\leq C_0$ and $K_0:=\sqrt{\frac{\rho}{2\pi
(1-\rho
)}}\leq\sqrt{\frac{\rho}{2\pi(1-C_0)}}$, we have
\begin{eqnarray*}
\varepsilon&\leq&\mathrm{e}^{t^2/2}\Biggl(1+\sum_{m=1}^{\infty}
\bigl(K_0^2d\bigr)^m\cdot m!2(2
\sqrt{2}et)^{2m}/(2m)^{2m}\Biggr)
\\
&\leq& \mathrm{e}^{t^2/2}\Biggl(1+\sum_{m=1}^{\infty}
m!2 \biggl(2\sqrt{\frac{d\rho
}{\pi
(1-C_0)}}et \biggr)^{2m}/(2m)^{2m}
\Biggr).
\end{eqnarray*}
Finally, we have for any $m\geq1$
\[
2m!\cdot m!\leq(2m)^{2m},
\]
implying that
%
\begin{equation}
\label{eq:new2} \varepsilon\leq \mathrm{e}^{t^2/2}\cdot\exp\bigl(4d\rho
\mathrm{e}^2/\pi\cdot t^2\bigr)=\exp \biggl\{ \biggl(
\frac{1}{2}+\frac{4d \rho \mathrm{e}^2}{\pi(1-C_0)} \biggr)t^2 \biggr\},
\end{equation}
where the term $\frac{1}{2}+\frac{4d \rho \mathrm{e}^2}{\pi(1-C_0)}$ is in the
same scale of $\|\Tb\|_2=1+(d-1)\cdot\frac{2}{\pi}\arcsin(\rho)$.
This completes the proof.
\end{pf*}

Corollary~\ref{cor:compound} can be proved similar to Theorem~\ref
{thm:example2}.

\begin{pf*}{Proof of Corollary~\ref{cor:compound}}
Letting $J_k=\{1+\sum_{j=1}^{k-1}d_j,\ldots,\sum_{j=1}^kd_j\}$. By the
product structure of the Gaussian distribution, we have
\[
\E\exp\bigl(t\bv^T\sign(\bX-\tilde{\bX})\bigr) = \prod
_{k=1}^q \E\exp \bigl(t\bv _{J_k}^T
\sign(\bX-\tilde{\bX})_{J_k}\bigr).
\]
Here we note that the bound in equation \eqref{eq:new2} also holds for
each $\E\exp(t\bv_{J_k}^T\sign(\bX-\tilde{\bX})_{J_k})$ by checking
equation \eqref{eq:new1}. Accordingly,
\begin{eqnarray*}
\prod_{k=1}^q\E\exp\bigl(t
\bv_{J_k}^T\sign(\bX-\tilde{\bX})_{J_k}\bigr) &
\leq& \prod_{k=1}^q\exp \biggl\{ \biggl(
\frac{1}{2}+\frac{4d_k \rho_k
\mathrm{e}^2}{\pi
(1-C_1)} \biggr)t^2 \biggr\}
\\
&\leq&\exp \biggl\{t^2 \biggl(\frac{q}{2}+\frac{4\mathrm{e}^2q}{\pi
(1-C_1)}
\max_k(d_k\rho_k) \biggr) \biggr\}.
\end{eqnarray*}
Because $q$ is upper bounded by a fixed constant, we have $\bv^T\sign
(\bX-\tilde{\bX})$ is sub-Gaussian. This completes the proof.
\end{pf*}

\subsection{Proof of Theorem \texorpdfstring{\protect\ref{thm:main}}{4.11}}\label{sec5.6}




%
\begin{pf}
We first prove that \eqref{eq:thmmain1} in Theorem~\ref{thm:main}
holds. Letting $\zeta:=K\|\Tb\|_2^2$, we aim to prove that with
probability larger than or equal to $1-2\alpha$,
%
\begin{eqnarray}
\label{eq:mainbound1} \sup_{\mathbf{b}\in\S^{s-1}}\sup_{J_{s}\in\{1,\ldots,d\}}\bigl
\llvert \mathbf {b}^{T}[\hat {\Tb}-\Tb]_{J_{s}, J_{s}}\mathbf{b}\bigr
\rrvert \leq2(8\zeta )^{1/2}\sqrt {\frac
{s(3+\log(d/s)) + \log(1/\alpha)}{n}}.
\end{eqnarray}

For the sphere $\S^{s-1}$ equipped with Euclidean metric, we let
$\mathcal{N}_{\varepsilon}$ be a subset of $\S^{s-1}$ such that for any
$\bv\in\S^{s-1}$, there exists $\mathbf{u}\in\mathcal
{N}_{\varepsilon}$ subject
to $\|\mathbf{u}-\bv\|_2\leq\varepsilon$. The cardinal number of
$\cN
_\varepsilon
$ has the upper bound
\[
\card(\mathcal{N}_{\varepsilon})< \biggl(1+\frac{2}{\varepsilon
}
\biggr)^{s}.
\]
Let $\mathcal{N}_{1/4}$ be a $(1/4)$-net of $\S^{s-1}$. Then the
cardinality of $\cN_{1/4}$
is bounded by $9^{s}$. Moreover, for any symmetric matrix $\Mb\in
\reals
^{s\times s}$,
\[
\sup_{\bv\in\S^{s-1}}\bigl|\bv^{T}\Mb\bv\bigr|\le\frac{1}{1-2\varepsilon
}\sup
_{\bv
\in\mathcal{N}_{\varepsilon}}\bigl|\bv^{T}\Mb\bv\bigr|.
\]
This implies that
\[
\sup_{\bv\in\S^{s-1}}\bigl|\bv^{T}\Mb\bv\bigr|\le2\sup
_{\bv\in\mathcal
{N}_{1/4}}\bigl|\bv^{T}\Mb\bv\bigr|.
\]
Let $\beta>0$ be a constant defined as
\[
\beta:=(8\zeta)^{1/2}\sqrt{\frac{s(3+\log(d/s)) + \log(1/\alpha)}{n}}.
\]
We have
\begin{eqnarray*}
&&\P \Bigl(\sup_{\mathbf{b}\in S^{s-1}}\sup_{J_{s}\subset\{1,\ldots
,d\}
}\bigl\llvert
\mathbf{b} ^{T} [\hat{\Tb}-\Tb ]_{J_{s}, J_{s}}\mathbf{b}\bigr\rrvert
>2\beta \Bigr)
\\
&&\quad\le \P \Bigl(\sup_{\mathbf{b}\in\mathcal{N}_{1/4}}\sup_{J_{s}\subset\{1,\ldots
,d\}}\bigl
\llvert \mathbf{b}^{T} [\hat{\Tb}-\Tb ]_{J_{s},
J_{s}}\mathbf{b}\bigr
\rrvert >\beta \Bigr)
\\
&&\quad\le 9^{s}\pmatrix{ d
\cr
s } \P \biggl( \bigl\llvert \mathbf{b}^{T}
[\hat{\Tb}-\Tb ]_{J_{s},
J_{s}}\mathbf{b}\bigr\rrvert >(8\zeta)^{1/2}
\sqrt{\frac{s(3+\log(d/s)) +
\log
(1/\alpha)}{n}},\\
&& \hspace*{66pt}\mbox{for fixed $\mathbf{b}$ and $J_{s}$ } \biggr).
\end{eqnarray*}
Thus, if we can show that for any fixed $\mathbf{b}$ and $J_{s}$ holds
%
\begin{equation}
\label{eq:boundE1} \P \bigl(\bigl\llvert \mathbf{b}^{T} [\hat{\Tb}-\Tb
]_{J_{s},
J_{s}}\mathbf{b}\bigr\rrvert >t \bigr) \le2\mathrm{e}^{-nt^{2}/(8\zeta)},
\end{equation}
then using the bound $
{d\choose
s}
<\{ed/(s)\}^{s}$, we have
\begin{eqnarray*}
&&9^{s}\pmatrix{ d
\cr
s }\P \biggl( \bigl\llvert \mathbf{b}^{T}
[\hat{\Tb}-\Tb ]_{J_{s},
J_{s}}\mathbf{b}\bigr\rrvert >(8\zeta)^{1/2}
\sqrt{\frac{s(3+\log(d/s)) +
\log
(1/\alpha)}{n}},
 \mbox{ for fixed $\mathbf{b}$ and $J$ } \biggr)
\\
&&\quad\le 2\exp\bigl\{s(1+\log9-\log s)+s\log d-s(3+\log d-\log s) -\log (1/\alpha)
\bigr\}
\\
&&\quad\le 2\alpha.
\end{eqnarray*}
It gives that with probability greater than $1-2\alpha$
the bound in equation (\ref{eq:mainbound1}) holds.

Finally, we show that equation \eqref{eq:boundE1} holds. For any $t$,
we have
\begin{eqnarray*}
&& \mathbb{E}\exp \bigl\{ t\cdot\mathbf{b}^{T} [\hat{\Tb}-\Tb
]_{J_{s}, J_{s}}\mathbf{b} \bigr\}
\\
&&\quad=  \mathbb{E}\exp \biggl\{ t\cdot\sum_{j\neq k\in
J_{s}}b_{j}b_{k}
(\hat{\tau}_{jk}-\tau_{jk} ) \biggr\}
\\
&&\quad=  \mathbb{E}\exp \biggl\{ t\cdot\frac{1}{{n\choose 2}}\sum
_{i<i'}\sum_{j\neq
k\in J_{s}}b_{j}b_{k}
\bigl(\sign\bigl((\mathbf{x}_{i}-\mathbf {x}_{i'})_{j}(
\mathbf {x}_{i}-\mathbf{x} _{i'})_{k}\bigr)-
\tau_{jk} \bigr) \biggr\}
.
\end{eqnarray*}
Let $S_n$ represent the permutation group of $\{1,\ldots,n\}$. For any
$\sigma\in S_n$, let $(i_1,\ldots,i_n):=\sigma(1,\ldots,n)$
represent a
permuted series of $\{1,\ldots,n\}$ and $\mathrm{O}(\sigma):=\{
(i_1,i_2),(i_3,i_4),\ldots,\break  (i_{n-1},i_n)\}$. In particular, we denote
$\mathrm{O}(\sigma_0):=\{(1,2),(3,4),\ldots,(n-1,n)\}$. By simple calculation,
%
\begin{eqnarray}
\label{eq:Ee^tbE1b} &&\hspace*{-6pt}\mathbb{E}\exp \biggl\{ t\cdot\frac{1}{{n\choose 2}}\sum
_{i<i'}\sum_{j\neq
k\in J_{s}}b_{j}b_{k}
\bigl(\sign\bigl((\mathbf{x}_{i}-\mathbf {x}_{i'})_{j}(
\mathbf {x}_{i}-\mathbf{x} _{i'})_{k}\bigr)-
\tau_{jk} \bigr) \biggr\}
\nonumber
\\
&&\hspace*{-8pt}\quad=  \mathbb{E}\exp \biggl\{ t\cdot\frac{1}{\card(S_n)}\sum
_{\sigma
\in
S_n} \frac{2}{n}\sum_{(i,i')\in \mathrm{O}(\sigma) }
\sum_{j\neq k\in
J_{s}}b_{j}b_{k} \bigl(
\sign\bigl((\mathbf{x}_{i}-\mathbf {x}_{i'})_{j}(
\mathbf {x}_{i}-\mathbf{x} _{i'})_{k}\bigr)-
\tau_{jk} \bigr) \biggr\}\qquad
\nonumber
\\[-8pt]
\\[-8pt]
\nonumber
&&\hspace*{-8pt}\quad\leq \frac{1}{\card(S_n)}\sum_{\sigma\in S_n} \mathbb{E}\exp
\biggl\{ t\cdot\frac{2}{n}\sum_{(i,i')\in \mathrm{O}(\sigma) }\sum
_{j\neq k\in
J_{s}}b_{j}b_{k} \bigl(\sign
\bigl((\mathbf{x}_{i}-\mathbf {x}_{i'})_{j}(
\mathbf {x}_{i}-\mathbf{x} _{i'})_{k}\bigr)-
\tau_{jk} \bigr) \biggr\}
\\
&&\hspace*{-8pt}\quad=  \mathbb{E}\exp \biggl\{ t\cdot\frac{2}{n}\sum
_{(i,i')\in
\mathrm{O}(\sigma
_0)}\sum_{j\neq k\in J_{s}}
b_{j}b_{k} \bigl(\sign\bigl((\mathbf {x}_{i}-
\mathbf{x} _{i'})_{j}(\mathbf{x}_{i}-
\mathbf{x}_{i'})_{k}\bigr)-\tau_{jk} \bigr)
\biggr\}.\nonumber
\end{eqnarray}
The inequality is due to the Jensen's inequality.

Let $m:=n/2$ and remind that $\bX=(X_1,\ldots,X_d)^T\sim \mathit{TE}_d(\bSigma
,\xi;f_1,\ldots,f_d)$. Let $\tilde{\bX}=(\tilde{X}_1,\ldots
,\tilde
{X}_d)^T$ be an independent copy of $\bX$. By equation \eqref
{eq:subgaussian}, we have that for any $|t|<t_0$ and $\bv\in\S^{d-1}$,
\[
\E\exp\bigl[t\bigl\{\bigl(\bv^T\sign(\bX-\tilde{\bX})
\bigr)^2-\E\bigl(\bv^T\sign(\bX -\tilde{\bX })
\bigr)^2\bigr\}\bigr] \leq \mathrm{e}^{ \zeta t^2}.
\]

In particular, letting $\bv_{J_{s}}=\mathbf{b}$ and $\bv
_{J_{s}^C}=\zero
$, we have
%
\begin{eqnarray}
\label{eq: LapTransf(Y)bound} \E\exp \biggl\{t\sum_{j\neq k\in J_{s}}
b_{j}b_{k} \bigl(\sign\bigl((\bX -\tilde {
\bX})_{j}(\bX-\tilde{\bX})_{k}\bigr)-\tau_{jk}
\bigr) \biggr\} \leq \mathrm{e}^{
\zeta t^2}.
\end{eqnarray}
Then we are able to continue equation \eqref{eq:Ee^tbE1b} as
%
\begin{eqnarray}
\label{eq:Ee^tbE1b2} &&\mathbb{E}\exp \biggl\{ t\cdot\frac{2}{n}\sum
_{(i,i')\in \mathrm{O}(\sigma
_0)}\sum_{j\neq k\in J_{s}}
b_{j}b_{k} \bigl(\sign\bigl((\mathbf{x}_{i}-
\mathbf {x}_{i'})_{j}(\mathbf{x} _{i}-
\mathbf{x}_{i'})_{k}\bigr)-\tau_{jk} \bigr)
\biggr\}
\nonumber
\\
&&\quad=  \mathbb{E}\exp \Biggl\{ \frac{t}{m}\sum_{i=1}^{m}
\biggl\{\sum_{j\neq
k\in J_{s}} b_{j}b_{k}
\bigl(\sign\bigl((\mathbf{x}_{2i}-\mathbf {x}_{2i-1})_{j}(
\mathbf{x}_{2i}-\mathbf{x} _{2i-1})_{k}\bigr)-
\tau_{jk} \bigr) \biggr\} \Biggr\}
\nonumber
\\[-8pt]
\\[-8pt]
\nonumber
&&\quad=  \bigl(\mathbb{E}\mathrm{e}^{({t}/{m}) (\sign((\bX-\tilde{\bX
})_{j}(\bX
-\tilde{\bX})_{k})-\tau_{jk} )} \bigr)^{m}
\\
&&\quad\le \mathrm{e}^{\zeta t^{2}/m},\nonumber
\end{eqnarray}
where by equation \eqref{eq:subgaussian}, the last inequality holds for
any $|t/m|<t_0$. Accordingly, choosing $t=\beta m/(2\zeta)$, by Markov
inequality, we have for sufficiently large $n$,
%
\begin{equation}
\label{eq:upperE1} \P \bigl(\mathbf{b}^{T} [\hat{\Tb}-\Tb
]_{J_{s},J_{s}}\mathbf {b}>\beta \bigr)\le \mathrm{e}^{-n\beta^{2}/(8\zeta)}
\qquad \mbox{for all }
\beta<2\zeta t_0.
\end{equation}
Because $t_0\|\Tb\|_2>C$ for some generic constant $C$, we have
$2\zeta t_0\geq2CK^{1/2}\zeta^{1/2}$, and hence as long as $\beta
\leq
2CK^{1/2}\zeta^{1/2}$, \eqref{eq:upperE1} holds.

By symmetry, we have the same bound for $\P (\mathbf{b}^{T}
[\hat{\Tb
}-\Tb ]_{J_{s}, J_{s}}\mathbf{b}<-\beta )$ as in equation
(\ref
{eq:upperE1}). Together they give us equation (\ref{eq:boundE1}). This
completes the proof of the first part.

Using \eqref{eq:thmmain1}, we can now proceed to the quantify the term
\[
\sup_{\bv\in\S^{d-1},\|\bv\|_0\leq s} \bigl|\bv^T(\hat{\bSigma }-\bSigma )\bv\bigr|.
\]
We aim to prove that, under the conditions in Theorem~\ref{thm:main},
we have with probability larger than or equal to $1-2\alpha-\alpha^2$,
%
\begin{eqnarray}
\label{eq:mainbound} &&\sup_{\mathbf{b}\in\S^{s-1}}\sup_{J_{s}\in\{1,\ldots,d\}}\bigl
\llvert \mathbf {b}^{T}[\hat {\bSigma}-\bSigma]_{J_{s}, J_{s}}
\mathbf{b}\bigr\rrvert
\nonumber
\\[-8pt]
\\[-8pt]
\nonumber
&&\quad \leq\pi^2(8\zeta)^{1/2}\sqrt{\frac{s(3+\log(d/s)) + \log
(1/\alpha)}{n}}+
\pi^2\cdot\frac{s\log(d/\alpha)}{n}.
\end{eqnarray}

Using a similar argument as in the proof of Theorem~\ref
{thm:lowd_spectral}, we let $\Eb_1,\Eb_2\in\reals^{d\times d}$,
satisfying that for $j\neq k$,
\begin{eqnarray*}
[\Eb_{1}]_{jk} & =& \cos \biggl(\frac{\pi}{2}
\tau_{jk} \biggr)\frac
{\pi
}{2}(\hat{\tau}_{jk}-
\tau_{jk}),
\\
{}[\Eb_{2}]_{jk} & = &-\frac{1}{2}\sin(
\theta_{jk}) \biggl(\frac{\pi
}{2} \biggr)^{2}(\hat{
\tau}_{jk}-\tau_{jk})^{2},
\end{eqnarray*}
where $\theta_{jk}$ lies between $\tau_{jk}$ and $\hat{\tau}_{jk}$. We
then have
\[
\hat{\bSigma}-\bSigma=\Eb_{1}+\Eb_{2}.
\]
Let the event $\Omega_{2}$ be defined as
\[
\Omega_{2}:= \biggl\{\exists1\le j\ne k\le d, \bigl|[\Eb_{2}]_{jk}\bigr|>
\pi ^{2}\frac{\log(d/\alpha)}{n} \biggr\}.
\]
Using the result in the proof of Theorem~\ref{thm:lowd_spectral}, we
have $\P(\Omega_2)\leq\alpha^2$.
Moreover, conditioning on $\Omega_2$, for any $J_{s}\in\{1,\ldots,d\}$
and $\mathbf{b}\in\S^{s-1}$,
%
\begin{eqnarray}
\label{eq:7} \bigl\llvert \mathbf{b}^T[\Eb_2]_{J_{s},J_{s}}
\mathbf{b}\bigr\rrvert &\leq& \sqrt{\sum_{j,k\in
J_{s}}
[\Eb_2]_{jk}^2} \cdot\|\mathbf{b}
\|_2^2
\nonumber
\\
&\leq &s\cdot\pi^2\cdot\frac{\log(d/\alpha)}{n}
\\
&=& \pi^2\cdot\frac{s\log(d/\alpha)}{n}.\nonumber
\end{eqnarray}

We then proceed to control the term $\llvert \mathbf{b}^T[\Eb
_1]_{J_{s},J_{s}}\mathbf{b}\rrvert $. Using a similar argument as
shown in
equation \eqref{eq:6}, for $\bY=(Y_1,\ldots,Y_d)^T\sim N_d(\zero
,\bSigma
)$, any symmetric matrix $\Mb\in\reals^{d\times d}$, $\Wb$ with
$\Wb
_{jk}=\frac{\pi}{2}\cos(\frac{\pi}{2}\tau_{jk})$ and $\bv\in\S^{d-1}$
with $\|\bv\|_0\leq q$, we have
\begin{eqnarray*}
\bigl| \bv^T \Mb\circ\Wb\bv\bigr| &\le&\frac{\pi^2}{4} \E \biggl( \biggl
\llvert \sum_{j,k} v_j v_k
\Mb_{jk} |Y_jY_k| \biggr\rrvert + \biggl
\llvert \sum_{j,k} v_j v_k
\Mb_{jk} Y_j Y_k \sign\bigl(Y_j'
Y_k'\bigr) \biggr\rrvert \biggr)
\nonumber
\\
&\le&\frac{\pi^2}{4} \sup_{\mathbf{b}\in\S^{d-1},\|\mathbf {b}\|
_0\leq q}\bigl|\mathbf{b}^T\Mb
\mathbf{b}\bigr| \cdot\E \biggl(2 \sum_j
v_j^2 Y_j^2 \biggr)
\nonumber
\\
&=& \frac{\pi^2}{4} \sup_{\mathbf{b}\in\S^{d-1},\|\mathbf{b}\|
_0\leq
q}\bigl|\mathbf{b}^T\Mb
\mathbf{b} \bigr| \cdot \biggl(2 \sum_j
v_j^2 \biggr)
\nonumber
\\
&=& \frac{\pi^2}{2} \sup_{\mathbf{b}\in\S^{d-1},\|\mathbf{b}\|
_0\leq
q}\bigl|\mathbf{b}^T\Mb
\mathbf{b}\bigr|.
\end{eqnarray*}
Accordingly, we have
\[
\sup_{\mathbf{b}\in\S^{s-1}}\sup_{J_{s}\in\{1,\ldots,d\}}\bigl\llvert \mathbf
{b}^T[\Eb _1]_{J_{s},J_{s}}\mathbf{b}\bigr\rrvert \leq
\frac{\pi^2}{2}\sup_{\mathbf{b}\in
\S
^{s-1}}\sup_{J_{s}\in\{1,\ldots,d\}}\bigl
\llvert \mathbf{b}^T[\hat{\Tb }-\Tb ]_{J_{s},J_{s}}\mathbf{b}\bigr
\rrvert .
\]
Combined with equations \eqref{eq:thmmain1}, \eqref{eq:7} and \eqref
{eq:T2lessS2}, we have the desired result in \eqref{eq:thmmain2}.

\end{pf}

\section{Discussions}\label{sec6}

This paper considers robust estimation of the correlation matrix using
the rank-based correlation coefficient estimator Kendall's tau and its
transformed version. We showed that the Kendall's tau is an very robust
estimator in high dimensions, in terms of that it can achieve the
parametric rate of convergence under various norms without any
assumption on the data distribution, and in particular, without
assuming any moment constraints. We further consider the
transelliptical family proposed in Han and Liu \cite{han2012transelliptical},
showing that a transformed version of the Kendall's tau attains the
parametric rate in estimating the latent Pearson's correlation matrix
without assuming any moment constraints. Moreover, unlike the Gaussian
case, the theoretical analysis performed here motivates new
understandings on rank-based estimators as well as new proof
techniques. These new understandings and proof techniques are of self-interest.

Han and Liu \cite{han2013dependent} studied the performance of the latent
generalized correlation matrix estimator on dependent data under some
mixing conditions and proved that $\hat\bSigma$ can attain a $s\sqrt
{\log d/(n\gamma)}$ rate of convergence under the restricted spectral
norm, where $\gamma\leq1$ reflects the impact of nonindependence on
the estimation accuracy. It is also interesting to consider extending
the results in this paper to dependent data under similar mixing
conditions and see whether a similar $\sqrt{s\log d/(n\gamma')}$ rate
of convergence can be attained. However, it is much more challenging to
obtain such results in dependent data. The current theoretical analysis
based on $U$-statistics is not sufficient to achieve this goal.

A problem closely related to the leading eigenvector estimation is
principal component detection, which is initiated in the work of
Berthet and Rigollet \cite
{berthet2012optimal,berthet2013computational}. It is interesting to
extend the analysis here to this setting and conduct sparse principal
component detection under the transelliptical family. It is worth
pointing out that Theorems \ref{thm:lowd_spectral} and \ref{thm:main}
in this paper can be exploited in measuring the statistical performance
of the corresponding detection of sparse principal components.

\begin{appendix}

\section*{Appendix}\label{app}

In this section, we provide a lemma quantifying the relationship
between Orlicz $\psi_{2}$-norm and the sub-Gaussian condition. Although
this result is well known, in order to quantify this relationship in
numbers, we include a proof here. We do not claim any original
contribution in this section.

\begin{lemma}\label{lem:subgaussian} For any random variable $Y\in
\reals
$, we say that $Y$ is a sub-Gaussian random variable with factor $c>0$
if and only if for any $t\in\reals$,
$\E\exp(tY)\leq\exp(ct^2)$. We than have $Y$ is sub-Gaussian if and
only if $\|Y\|_{\psi_2}$ is bounded. In particular, we have that if
$Y$ is sub-Gaussian with factor $c$, then
$
\|Y\|_{\psi_2}\leq\sqrt{12c}$.
If $\|Y\|_{\psi_2}\leq D\leq\infty$, then $Y$ is sub-Gaussian with
factor $c=5D^2/2$.
\end{lemma}
\begin{pf}
If $Y$ is sub-Gaussian, then for any $m>0$, we have
\begin{eqnarray*}
\E\exp\bigl(|Y/m|^2\bigr)&=&1+\int_{0}^{\infty}
\P \biggl(\frac{Y^2}{m^2}>t \biggr)\mathrm{e}^t\,\mathrm{d}t
\\
&=&1+\int_0^{\infty}\P\bigl(|Y|>m\sqrt{t}\bigr)\mathrm{e}^t\,\mathrm{d}t.
\end{eqnarray*}
By Markov inequality, we know that if $Y$ is sub-Gaussian, then for any $t>0$
\[
\P\bigl(|Y|>t\bigr)\leq2\exp\bigl(-t^2/(4c)\bigr).
\]
Accordingly, we can proceed the proof
\begin{eqnarray*}
\E\exp\bigl(|Y/m|^2\bigr) &\leq&1+2\int_0^{\infty}\mathrm{e}^{-m^2t/(4c)}
\cdot \mathrm{e}^t\,\mathrm{d}t
\\
&=& 1+2\int_0^{\infty}\mathrm{e}^{-(m^2/(4c)-1)t}\,\mathrm{d}t
\\
&=&1+\frac{2}{m^2/(4c)-1}.
\end{eqnarray*}
Picking $m=\sqrt{12c}$, we have $\E\exp(|Y/m|^2)\leq2$. Accordingly,
$\|Y\|_{\psi_2}\leq\sqrt{12c}$.
On the other hand, if $\|Y\|_{\psi_2}\leq\infty$, then there exists
some $m<\infty$ such that $\E\exp(|Y/m|^2)\leq2$. Using integration by
part, it is easy to check that
\[
\exp(a)=1+a^2\int_0^1(1-y)\mathrm{e}^{ay}\,\mathrm{d}y.
\]
This implies that
\begin{eqnarray*}
\E\exp(tX) &=& 1+\int_0^1(1-u)\E
\bigl[(tX)^2\exp(utX)\bigr]\,\mathrm{d}u
\\
&\leq&1+ t^2\E\bigl(X^2\exp\bigl(|tX|\bigr)\bigr)\int
_0^1(1-u)\,\mathrm{d}u
\\
&\leq&1+ \frac{t^2}{2}\E\bigl(X^2\mathrm{e}^{|tX|}\bigr).
\end{eqnarray*}
Using the fact that for any $a,b\in\reals$, $|ab|\leq\frac{a^2+b^2}{2}$
and $a\leq \mathrm{e}^a$, we can further prove that
\begin{eqnarray*}
\E\exp(tX) &\leq&1+ \frac{t^2}{2}\E\bigl(X^2\mathrm{e}^{|tX|}
\bigr)
\\
&\leq&1+m^2t^2\mathrm{e}^{m^2t^2/2}\E \biggl(
\frac
{X^2}{2m^2}\mathrm{e}^{X^2/(2m^2)} \biggr)
\\
&\leq&1+m^2t^2\mathrm{e}^{m^2t^2/2}\E \mathrm{e}^{X^2/m^2}
\\
&\leq&\bigl(1+2m^2t^2\bigr)\mathrm{e}^{m^2t^2/2}
\\
&\le& \mathrm{e}^{5m^2t^2/2}.
\end{eqnarray*}
The last inequality is due to the fact that for any $a\in\reals$,
$1+a\leq \mathrm{e}^a$. Accordingly, $X$ is sub-Gaussian with the factor $c= 5m^2/2$.
\end{pf}
\end{appendix}

\section*{Acknowledgement}
We sincerely thank Marten Wegkamp for his very
helpful discussions and generously providing
independent credit for our work. We thank the
Editor, Associate Editor, and two anonymous
referees for their very valuable comments, which
significantly improve the quality of our work. We also
thank Xiuyuan Cheng, Ramon van Handel, Philippe
Rigollet, and Luo Xiao for their many helps. Fang
Han's research was supported by NIBIB-EB012547.
Han Liu's research was supported by the NSF
CAREER Award DMS-1454377, NSF IIS-1546482,
NSF IIS-1408910, NSF IIS-1332109, NIH
R01-MH102339, NIH R01-GM083084, and NIH
R01-HG06841.


\begin{thebibliography}{37}

\bibitem{baik2006eigenvalues}
%
\begin{barticle}[mr]
\bauthor{\bsnm{Baik},~\bfnm{Jinho}\binits{J.}} \AND
\bauthor{\bsnm{Silverstein},~\bfnm{Jack~W.}\binits{J.W.}}
(\byear{2006}).
\btitle{Eigenvalues of large sample covariance matrices of spiked
population models}.
\bjournal{J. Multivariate Anal.}
\bvolume{97}
\bpages{1382--1408}.
\bid{doi={10.1016/j.jmva.2005.08.003}, issn={0047-259X}, mr={2279680}}
\end{barticle}
%
\bptok{imsref}%
\endbibitem

\bibitem{berthet2013computational}
%
\begin{bmisc}[auto:parserefs-M02]
\bauthor{\bsnm{Berthet},~\bfnm{Q.}\binits{Q.}} \AND
\bauthor{\bsnm{Rigollet},~\bfnm{P.}\binits{P.}}
(\byear{2013}).
\bhowpublished{Computational lower bounds for sparse PCA.
Preprint. Available at
\arxivurl{arXiv:1304.0828}}.
\end{bmisc}
%
\bptok{imsref}%
\endbibitem

\bibitem{berthet2012optimal}
%
\begin{barticle}[mr]
\bauthor{\bsnm{Berthet},~\bfnm{Quentin}\binits{Q.}} \AND
\bauthor{\bsnm{Rigollet},~\bfnm{Philippe}\binits{P.}}
(\byear{2013}).
\btitle{Optimal detection of sparse principal components in high dimension}.
\bjournal{Ann. Statist.}
\bvolume{41}
\bpages{1780--1815}.
\bid{doi={10.1214/13-AOS1127}, issn={0090-5364}, mr={3127849}}
\end{barticle}
%
\bptok{imsref}%
\endbibitem

\bibitem{bickel2008regularized}
%
\begin{barticle}[mr]
\bauthor{\bsnm{Bickel},~\bfnm{Peter~J.}\binits{P.J.}} \AND
\bauthor{\bsnm{Levina},~\bfnm{Elizaveta}\binits{E.}}
(\byear{2008}).
\btitle{Regularized estimation of large covariance matrices}.
\bjournal{Ann. Statist.}
\bvolume{36}
\bpages{199--227}.
\bid{doi={10.1214/009053607000000758}, issn={0090-5364}, mr={2387969}}
\end{barticle}
%
\bptok{imsref}%
\endbibitem

\bibitem{bickel2008covariance}
%
\begin{barticle}[mr]
\bauthor{\bsnm{Bickel},~\bfnm{Peter~J.}\binits{P.J.}} \AND
\bauthor{\bsnm{Levina},~\bfnm{Elizaveta}\binits{E.}}
(\byear{2008}).
\btitle{Covariance regularization by thresholding}.
\bjournal{Ann. Statist.}
\bvolume{36}
\bpages{2577--2604}.
\bid{doi={10.1214/08-AOS600}, issn={0090-5364}, mr={2485008}}
\end{barticle}
%
\bptok{imsref}%
\endbibitem

\bibitem{boente2012characterization}
%
\begin{btechreport}[auto:parserefs-M02]
\bauthor{\bsnm{Boente},~\bfnm{G.}\binits{G.}},
\bauthor{\bsnm{Barrerab},~\bfnm{M.~S.}\binits{M.S.}} \AND
\bauthor{\bsnm{Tylerc},~\bfnm{D.~E.}\binits{D.E.}}
(\byear{2012}).
\btitle{A characterization of elliptical distributions and some
optimality properties of principal components for functional data}.
\btype{Technical report}.
\bnote{Available at \surl{http://www.stat.ubc.ca/\textasciitilde
matias/Property\_FPCA\_rev1.pdf}}.
\end{btechreport}
%
\bptok{imsref}%
\endbibitem

\bibitem{bunea2012sample}
%
\begin{barticle}[mr]
\bauthor{\bsnm{Bunea},~\bfnm{Florentina}\binits{F.}} \AND
\bauthor{\bsnm{Xiao},~\bfnm{Luo}\binits{L.}}
(\byear{2015}).
\btitle{On the sample covariance matrix estimator of reduced effective
rank population matrices, with applications to f{PCA}}.
\bjournal{Bernoulli}
\bvolume{21}
\bpages{1200--1230}.
\bid{doi={10.3150/14-BEJ602}, issn={1350-7265}, mr={3338661}}
\bptnote{check volume, check pages, check year}%
\end{barticle}
%
\bptok{imsref}%
\endbibitem

\bibitem{cai2013optimal}
%
\begin{barticle}[mr]
\bauthor{\bsnm{Cai},~\bfnm{Tony}\binits{T.}},
\bauthor{\bsnm{Ma},~\bfnm{Zongming}\binits{Z.}} \AND
\bauthor{\bsnm{Wu},~\bfnm{Yihong}\binits{Y.}}
(\byear{2015}).
\btitle{Optimal estimation and rank detection for sparse spiked
covariance matrices}.
\bjournal{Probab. Theory Related Fields}
\bvolume{161}
\bpages{781--815}.
\bid{doi={10.1007/s00440-014-0562-z}, issn={0178-8051}, mr={3334281}}
\bptnote{check volume, check pages, check year}%
\end{barticle}
%
\bptok{imsref}%
\endbibitem

\bibitem{cai2010optimal}
%
\begin{barticle}[mr]
\bauthor{\bsnm{Cai},~\bfnm{T.~Tony}\binits{T.T.}},
\bauthor{\bsnm{Zhang},~\bfnm{Cun-Hui}\binits{C.-H.}} \AND
\bauthor{\bsnm{Zhou},~\bfnm{Harrison~H.}\binits{H.H.}}
(\byear{2010}).
\btitle{Optimal rates of convergence for covariance matrix estimation}.
\bjournal{Ann. Statist.}
\bvolume{38}
\bpages{2118--2144}.
\bid{doi={10.1214/09-AOS752}, issn={0090-5364}, mr={2676885}}
\end{barticle}
%
\bptok{imsref}%
\endbibitem

\bibitem{cai2012minimax}
%
\begin{barticle}[mr]
\bauthor{\bsnm{Cai},~\bfnm{T.~Tony}\binits{T.T.}} \AND
\bauthor{\bsnm{Zhou},~\bfnm{Harrison~H.}\binits{H.H.}}
(\byear{2012}).
\btitle{Minimax estimation of large covariance matrices under {$\ell
_1$}-norm}.
\bjournal{Statist. Sinica}
\bvolume{22}
\bpages{1319--1349}.
\bid{issn={1017-0405}, mr={3027084}}
\bptnote{check pages}%
\end{barticle}
%
\bptok{imsref}%
\endbibitem

\bibitem{chung2006complex}
%
\begin{bbook}[mr]
\bauthor{\bsnm{Chung},~\bfnm{Fan}\binits{F.}} \AND
\bauthor{\bsnm{Lu},~\bfnm{Linyuan}\binits{L.}}
(\byear{2006}).
\btitle{Complex Graphs and Networks}.
\bseries{CBMS Regional Conference Series in Mathematics}
\bvolume{107}.
\blocation{Providence, RI}:
\bpublisher{Amer. Math. Soc}.
\bid{mr={2248695}}
\end{bbook}
%
\bptok{imsref}%
\endbibitem

\bibitem{embrechts2003modelling}
%
\begin{barticle}[auto:parserefs-M02]
\bauthor{\bsnm{Embrechts},~\bfnm{P.}\binits{P.}},
\bauthor{\bsnm{Lindskog},~\bfnm{F.}\binits{F.}} \AND
\bauthor{\bsnm{McNeil},~\bfnm{A.}\binits{A.}}
(\byear{2003}).
\btitle{Modelling dependence with copulas and applications to risk management}.
\bjournal{Handbook of Heavy Tailed Distributions in Finance}
\bvolume{8}
\bpages{329--384}.
\end{barticle}
%
\bptok{imsref}%
\endbibitem

\bibitem{fang2002}
%
\begin{barticle}[mr]
\bauthor{\bsnm{Fang},~\bfnm{Hong-Bin}\binits{H.-B.}},
\bauthor{\bsnm{Fang},~\bfnm{Kai-Tai}\binits{K.-T.}} \AND
\bauthor{\bsnm{Kotz},~\bfnm{Samuel}\binits{S.}}
(\byear{2002}).
\btitle{The meta-elliptical distributions with given marginals}.
\bjournal{J. Multivariate Anal.}
\bvolume{82}
\bpages{1--16}.
\bid{doi={10.1006/jmva.2001.2017}, issn={0047-259X}, mr={1918612}}
\end{barticle}
%
\bptok{imsref}%
\endbibitem

\bibitem{fangsymmetric}
%
\begin{bbook}[mr]
\bauthor{\bsnm{Fang},~\bfnm{Kai~Tai}\binits{K.T.}},
\bauthor{\bsnm{Kotz},~\bfnm{Samuel}\binits{S.}} \AND
\bauthor{\bsnm{Ng},~\bfnm{Kai~Wang}\binits{K.W.}}
(\byear{1990}).
\btitle{Symmetric Multivariate and Related Distributions}.
\bseries{Monographs on Statistics and Applied Probability}
\bvolume{36}.
\blocation{London}:
\bpublisher{Chapman \& Hall}.
\bid{doi={10.1007/978-1-4899-2937-2}, mr={1071174}}
\end{bbook}
%
\bptok{imsref}%
\endbibitem

\bibitem{han2013dependent}
%
\begin{barticle}[auto:parserefs-M02]
\bauthor{\bsnm{Han},~\bfnm{F.}\binits{F.}} \AND
\bauthor{\bsnm{Liu},~\bfnm{H.}\binits{H.}}
(\byear{2013}).
\btitle{Principal component analysis on non-{G}aussian dependent data}.
\bjournal{J. Mach. Learn. Res. Workshop Conf. Proc.}
\bvolume{28}
\bpages{240--248}.
\end{barticle}
%
\bptok{imsref}%
\endbibitem

\bibitem{han2012semiparametric}
%
\begin{barticle}[auto:parserefs-M02]
\bauthor{\bsnm{Han},~\bfnm{F.}\binits{F.}} \AND
\bauthor{\bsnm{Liu},~\bfnm{H.}\binits{H.}}
(\byear{2014}).
\btitle{High dimensional semiparametric scale-invariant principal
component analysis}.
\bjournal{IEEE Trans. Pattern Anal. Mach. Intell.}
\bvolume{36}
\bpages{2016--2032}.
\end{barticle}
%
\bptok{imsref}%
\endbibitem

\bibitem{han2012transelliptical}
%
\begin{barticle}[auto:parserefs-M02]
\bauthor{\bsnm{Han},~\bfnm{F.}\binits{F.}} \AND
\bauthor{\bsnm{Liu},~\bfnm{H.}\binits{H.}}
(\byear{2014}).
\btitle{Scale-invariant sparse {PCA} on high dimensional meta-elliptical
data}.
\bjournal{J.~Am. Stat. Assoc.}
\bvolume{109}
\bpages{275--287}.
\bid{mr={3180563}}
\end{barticle}
%
\bptok{imsref}%
\endbibitem

\bibitem{han2013coda}
%
\begin{barticle}[mr]
\bauthor{\bsnm{Han},~\bfnm{Fang}\binits{F.}},
\bauthor{\bsnm{Zhao},~\bfnm{Tuo}\binits{T.}} \AND
\bauthor{\bsnm{Liu},~\bfnm{Han}\binits{H.}}
(\byear{2013}).
\btitle{C{ODA}: High dimensional copula discriminant analysis}.
\bjournal{J. Mach. Learn. Res.}
\bvolume{14}
\bpages{629--671}.
\bid{issn={1532-4435}, mr={3033343}}
\end{barticle}
%
\bptok{imsref}%
\endbibitem

\bibitem{hoeffding1963probability}
%
\begin{barticle}[mr]
\bauthor{\bsnm{Hoeffding},~\bfnm{Wassily}\binits{W.}}
(\byear{1963}).
\btitle{Probability inequalities for sums of bounded random variables}.
\bjournal{J. Amer. Statist. Assoc.}
\bvolume{58}
\bpages{13--30}.
\bid{issn={0162-1459}, mr={0144363}}
\end{barticle}
%
\bptok{imsref}%
\endbibitem

\bibitem{hogg1994introduction}
%
\begin{bbook}[auto:parserefs-M02]
\bauthor{\bsnm{Hogg},~\bfnm{R.~V.}\binits{R.V.}} \AND
\bauthor{\bsnm{Craig},~\bfnm{A.}\binits{A.}}
(\byear{2012}).
\btitle{Introduction to Mathematical Statistics},
\bedition{7th} ed.
\blocation{Upper Saddle River}:
\bpublisher{Harlow, Essex}.
\end{bbook}
%
\bptok{imsref}%
\endbibitem

\bibitem{hubbard1959calculation}
%
\begin{barticle}[auto:parserefs-M02]
\bauthor{\bsnm{Hubbard},~\bfnm{J.}\binits{J.}}
(\byear{1959}).
\btitle{Calculation of partition functions}.
\bjournal{Phys. Rev. Lett.}
\bvolume{3}
\bpages{77}.
\end{barticle}
%
\bptok{imsref}%
\endbibitem

\bibitem{johnson1990matrix}
%
\begin{bbook}[mr]
\beditor{\bsnm{Johnson},~\bfnm{Charles R.}\binits{C.R.}}, ed.
(\byear{1990}).
\btitle{Matrix Theory and Applications}.
\bseries{Proceedings of Symposia in Applied Mathematics}
\bvolume{40}.
\blocation{Providence, RI}:
\bpublisher{Amer. Math. Soc.}
\bid{doi={10.1090/psapm/040}, mr={1059481}}
\end{bbook}
%
\bptok{imsref}%
\endbibitem

\bibitem{johnstone2001distribution}
%
\begin{barticle}[mr]
\bauthor{\bsnm{Johnstone},~\bfnm{Iain~M.}\binits{I.M.}}
(\byear{2001}).
\btitle{On the distribution of the largest eigenvalue in principal
components analysis}.
\bjournal{Ann. Statist.}
\bvolume{29}
\bpages{295--327}.
\bid{doi={10.1214/aos/1009210544}, issn={0090-5364}, mr={1863961}}
\end{barticle}
%
\bptok{imsref}%
\endbibitem

\bibitem{jung2009pca}
%
\begin{barticle}[mr]
\bauthor{\bsnm{Jung},~\bfnm{Sungkyu}\binits{S.}} \AND
\bauthor{\bsnm{Marron},~\bfnm{J.~S.}\binits{J.S.}}
(\byear{2009}).
\btitle{P{CA} consistency in high dimension, low sample size context}.
\bjournal{Ann. Statist.}
\bvolume{37}
\bpages{4104--4130}.
\bid{doi={10.1214/09-AOS709}, issn={0090-5364}, mr={2572454}}
\end{barticle}
%
\bptok{imsref}%
\endbibitem

\bibitem{ledoux2001concentration}
%
\begin{bbook}[mr]
\bauthor{\bsnm{Ledoux},~\bfnm{Michel}\binits{M.}}
(\byear{2001}).
\btitle{The Concentration of Measure Phenomenon}.
\bseries{Mathematical Surveys and Monographs}
\bvolume{89}.
\blocation{Providence, RI}:
\bpublisher{Amer. Math. Soc.}
\bid{mr={1849347}}
\end{bbook}
%
\bptok{imsref}%
\endbibitem

\bibitem{lindskog2003kendall}
%
\begin{barticle}[auto:parserefs-M02]
\bauthor{\bsnm{Lindskog},~\bfnm{F.}\binits{F.}},
\bauthor{\bsnm{McNeil},~\bfnm{A.}\binits{A.}} \AND
\bauthor{\bsnm{Schmock},~\bfnm{U.}\binits{U.}}
(\byear{2003}).
\btitle{Kendall's tau for elliptical distributions}.
\bjournal{Credit risk: Measurement, Evaluation and Management}
\bpages{149--156}.
\end{barticle}
%
\bptok{imsref}%
\endbibitem

\bibitem{liu2012high}
%
\begin{barticle}[mr]
\bauthor{\bsnm{Liu},~\bfnm{Han}\binits{H.}},
\bauthor{\bsnm{Han},~\bfnm{Fang}\binits{F.}},
\bauthor{\bsnm{Yuan},~\bfnm{Ming}\binits{M.}},
\bauthor{\bsnm{Lafferty},~\bfnm{John}\binits{J.}} \AND
\bauthor{\bsnm{Wasserman},~\bfnm{Larry}\binits{L.}}
(\byear{2012}).
\btitle{High-dimensional semiparametric {G}aussian copula graphical models}.
\bjournal{Ann. Statist.}
\bvolume{40}
\bpages{2293--2326}.
\bid{doi={10.1214/12-AOS1037}, issn={0090-5364}, mr={3059084}}
\end{barticle}
%
\bptok{imsref}%
\endbibitem

\bibitem{liu2012transelliptical}
%
\begin{binproceedings}[auto:parserefs-M02]
\bauthor{\bsnm{Liu},~\bfnm{H.}\binits{H.}},
\bauthor{\bsnm{Han},~\bfnm{F.}\binits{F.}} \AND
\bauthor{\bsnm{Zhang},~\bfnm{C.-H.}\binits{C.-H.}}
(\byear{2012}).
\btitle{Transelliptical graphical models}.
In \bbooktitle{Proceedings of the Twenty-Fifth Annual Conference on
Neural Information Processing Systems}
\bpages{809--817}.
\end{binproceedings}
%
\bptok{imsref}%
\endbibitem

\bibitem{liu2009nonparanormal}
%
\begin{barticle}[mr]
\bauthor{\bsnm{Liu},~\bfnm{Han}\binits{H.}},
\bauthor{\bsnm{Lafferty},~\bfnm{John}\binits{J.}} \AND
\bauthor{\bsnm{Wasserman},~\bfnm{Larry}\binits{L.}}
(\byear{2009}).
\btitle{The nonparanormal: Semiparametric estimation of high
dimensional undirected graphs}.
\bjournal{J. Mach. Learn. Res.}
\bvolume{10}
\bpages{2295--2328}.
\bid{issn={1532-4435}, mr={2563983}}
\end{barticle}
%
\bptok{imsref}%
\endbibitem

\bibitem{lounici2012high}
%
\begin{barticle}[mr]
\bauthor{\bsnm{Lounici},~\bfnm{Karim}\binits{K.}}
(\byear{2014}).
\btitle{High-dimensional covariance matrix estimation with missing
observations}.
\bjournal{Bernoulli}
\bvolume{20}
\bpages{1029--1058}.
\bid{doi={10.3150/12-BEJ487}, issn={1350-7265}, mr={3217437}}
\bptnote{check pages, check year}%
\end{barticle}
%
\bptok{imsref}%
\endbibitem

\bibitem{tropp2011user}
%
\begin{barticle}[mr]
\bauthor{\bsnm{Tropp},~\bfnm{Joel~A.}\binits{J.A.}}
(\byear{2012}).
\btitle{User-friendly tail bounds for sums of random matrices}.
\bjournal{Found. Comput. Math.}
\bvolume{12}
\bpages{389--434}.
\bid{doi={10.1007/s10208-011-9099-z}, issn={1615-3375}, mr={2946459}}
\bptnote{check volume, check pages, check year}%
\end{barticle}
%
\bptok{imsref}%
\endbibitem

\bibitem{van2011bernstein}
%
\begin{barticle}[mr]
\bauthor{\bsnm{van~de Geer},~\bfnm{Sara}\binits{S.}} \AND
\bauthor{\bsnm{Lederer},~\bfnm{Johannes}\binits{J.}}
(\byear{2013}).
\btitle{The {B}ernstein--{O}rlicz norm and deviation inequalities}.
\bjournal{Probab. Theory Related Fields}
\bvolume{157}
\bpages{225--250}.
\bid{doi={10.1007/s00440-012-0455-y}, issn={0178-8051}, mr={3101846}}
\end{barticle}
%
\bptok{imsref}%
\endbibitem

\bibitem{vershynin2010introduction}
%
\begin{bincollection}[mr]
\bauthor{\bsnm{Vershynin},~\bfnm{Roman}\binits{R.}}
(\byear{2012}).
\btitle{Introduction to the non-asymptotic analysis of random matrices}.
In \bbooktitle{Compressed Sensing}
\bpages{210--268}.
\blocation{Cambridge}:
\bpublisher{Cambridge Univ. Press}.
\bid{mr={2963170}}
\bptnote{check pages, check year}%
\end{bincollection}
%
\bptok{imsref}%
\endbibitem

\bibitem{vu2012minimax}
%
\begin{barticle}[auto:parserefs-M02]
\bauthor{\bsnm{Vu},~\bfnm{V.}\binits{V.}} \AND
\bauthor{\bsnm{Lei},~\bfnm{J.}\binits{J.}}
(\byear{2012}).
\btitle{Minimax rates of estimation for sparse PCA in high dimensions}.
\bjournal{J. Mach. Learn. Res. Workshop Conf. Proc.}
\bvolume{22}
\bpages{1278--1286}.
\end{barticle}
%
\bptok{imsref}%
\endbibitem

\bibitem{marten2013}
%
\begin{bmisc}[auto:parserefs-M02]
\bauthor{\bsnm{Wegkamp},~\bfnm{M.}\binits{M.}} \AND
\bauthor{\bsnm{Zhao},~\bfnm{Y.}\binits{Y.}}
(\byear{2013}).
\bhowpublished{Analysis of elliptical copula correlation factor model
with Kendall's tau.
Personal communication.}
\end{bmisc}
%
\bptok{imsref}%
\endbibitem

\bibitem{xue2012regularized}
%
\begin{barticle}[mr]
\bauthor{\bsnm{Xue},~\bfnm{Lingzhou}\binits{L.}} \AND
\bauthor{\bsnm{Zou},~\bfnm{Hui}\binits{H.}}
(\byear{2012}).
\btitle{Regularized rank-based estimation of high-dimensional
nonparanormal graphical models}.
\bjournal{Ann. Statist.}
\bvolume{40}
\bpages{2541--2571}.
\bid{doi={10.1214/12-AOS1041}, issn={0090-5364}, mr={3097612}}
\end{barticle}
%
\bptok{imsref}%
\endbibitem

\bibitem{yuan2011truncated}
%
\begin{barticle}[mr]
\bauthor{\bsnm{Yuan},~\bfnm{Xiao-Tong}\binits{X.-T.}} \AND
\bauthor{\bsnm{Zhang},~\bfnm{Tong}\binits{T.}}
(\byear{2013}).
\btitle{Truncated power method for sparse eigenvalue problems}.
\bjournal{J. Mach. Learn. Res.}
\bvolume{14}
\bpages{899--925}.
\bid{issn={1532-4435}, mr={3063614}}
\end{barticle}
%
\bptok{imsref}%
\endbibitem

\end{thebibliography}
%
%





\printhistory
\end{document}